\pdfoutput=1

\documentclass[11pt]{article}

\usepackage[]{EMNLP2023}

\usepackage{times}
\usepackage{tabu}
\usepackage{latexsym}
\usepackage{stmaryrd}
\usepackage{amsmath}
\usepackage{multirow}
\usepackage{bbm}
\usepackage{graphicx}
\usepackage{amssymb}
\usepackage{booktabs}
\usepackage{adjustbox}
\usepackage{xspace}
\usepackage{algpseudocode}
\usepackage{algorithm}
\usepackage{graphicx}
\usepackage{caption}
\usepackage{subcaption}
\usepackage{tabularx}
\usepackage{colortbl}
\usepackage{xcolor}
\usepackage{tablefootnote}
\usepackage{enumitem}  
\usepackage{lipsum}
\usepackage{xspace}
\usepackage{bbding}
\usepackage{pifont}
\usepackage{arydshln}
\usepackage{array}
\newcommand{\PreserveBackslash}[1]{\let\temp=\\#1\let\\=\temp}
\newcolumntype{C}[1]{>{\PreserveBackslash\centering}p{#1}}
\newcolumntype{R}[1]{>{\PreserveBackslash\raggedleft}p{#1}}
\newcolumntype{L}[1]{>{\PreserveBackslash\raggedright}p{#1}}

\usepackage{hyperref}

\def\xHyphenate#1#2\wholeString {\if#1$%
    \else\transform{#1}%
    \takeTheRest#2\ofTheString\fi}
\def\takeTheRest#1\ofTheString\fi
{\fi \xHyphenate#1\wholeString}
\def\transform#1{\url{#1}\hskip 0pt plus 1pt}

\usepackage[T1]{fontenc}

\usepackage[utf8]{inputenc}

\usepackage{microtype}

\newcommand{\ex}[1]{\textit{#1}\xspace} 
\newcommand{\gl}[1]{``#1''\xspace}

\newcommand{\xmark}{\ding{55}}
\newcommand{\cmark}{\ding{51}}
\definecolor{mygreen}{RGB}{217, 234, 211}
\definecolor{myred}{RGB}{244, 204, 204}

\newcommand{\ok}{\cellcolor{mygreen}\cmark}
\newcommand{\no}{\cellcolor{myred}\xmark}

\usepackage{inconsolata}

\setlist{topsep=1pt,itemsep=1pt,partopsep=1pt, parsep=1pt}

\usepackage{xcolor}

\title{Zero-shot Sentiment Analysis in Low-Resource Languages Using a Multilingual Sentiment Lexicon}

\author{Fajri Koto$^{1}$ \qquad  Tilman Beck$^{2}$ \qquad Zeerak Talat$^{1}$ \\ \textbf{Iryna Gurevych}$^{1}$ \qquad \textbf{Timothy Baldwin}$^{1,3}$\\
	$^{1}$Department Natural Language Processing, MBZUAI \\
        $^{2}$Ubiquitous Knowledge Processing Lab, Technical University of Darmstadt \\
        $^{3}$The University of Melbourne\\
	\texttt{\small fajri.koto@mbzuai.ac.ae 
	} \\
}

\begin{document}
\maketitle
\begin{abstract}

Improving multilingual language models capabilities in low-resource languages is generally difficult due to the scarcity of large-scale data in those languages. 
In this paper, we relax the reliance on texts in low-resource languages by using multilingual lexicons in pretraining to enhance multilingual capabilities. Specifically, we focus on zero-shot sentiment analysis tasks across 34 languages, including 6 high/medium-resource languages, 25 low-resource languages, and 3 code-switching datasets. We demonstrate that pretraining using multilingual lexicons, without using
any sentence-level sentiment data, achieves superior zero-shot performance compared to models fine-tuned on English sentiment datasets, and large language models like GPT--3.5, BLOOMZ, and XGLM. These findings are observable for unseen low-resource languages to code-mixed scenarios involving high-resource languages.\footnote{Code and dataset can be found at: \url{https://github.com/fajri91/ZeroShotMultilingualSentiment}}
\end{abstract}

\section{Introduction}

When it comes to under-represented languages, multilingual language models~\cite{conneau-etal-2020-unsupervised,xue-etal-2021-mt5,devlin-etal-2019-bert,liu-etal-2020-multilingual-denoising} are often considered the most viable option in the current era of pretraining and fine-tuning, primarily due to the scarcity of labeled and unlabeled training data. However, the limited language coverage of these models often results in poor cross-lingual transfer to under-represented languages~\cite{xia-etal-2021-metaxl,wang-etal-2022-expanding}. 

Prior work has extended multilingual models~\cite{conneau-etal-2020-unsupervised,xue-etal-2021-mt5} to 
other languages by language-adaptive pretraining (i.e.,~continuing to pretrain on monolingual text) \cite[e.g.,][]{wang-etal-2020-extending,chau-etal-2020-parsing} and leveraging adapters \cite{pfeiffer-etal-2020-mad}. However, these language adaptation techniques are not compatible with low-resource languages due to the unavailability of adequate unlabeled monolingual texts. 

Lexicons are more readily accessible and offer broader language coverage than monolingual corpora like Wikipedia and the Bible, making them a promising resource for extending multilingual models to under-represented languages. This is because when studying a new language, a lexicon is generally the first resource that 
field linguists develop
to document its morpho-phonemics and basic vocabulary. Of the 7,000+ languages spoken worldwide, lexicons are available for approximately 70\% of them, while mBERT, Wikipedia/CommonCrawl, and the Bible are available for only 1\%, 4\%, and 23\%, respectively \cite{wang-etal-2022-expanding}. 

\begin{figure}[t]
    \centering
    \includegraphics[width=0.9\linewidth]{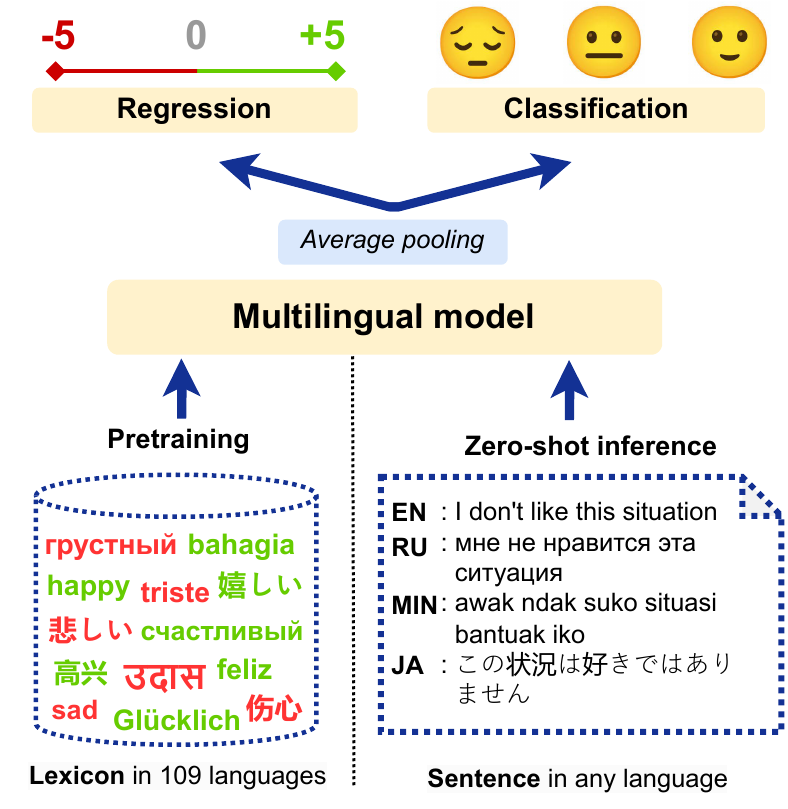} 
    \caption{\textbf{Left}: pretraining with a multilingual sentiment lexicon. \textbf{Right}: zero-shot inference using sentences or documents.}
    \label{fig:overview}
     \vspace{-0.35cm}
\end{figure}

In prior work,  \citet{wang-etal-2022-expanding} proposed to use the Panlex translation lexicon \cite{baldwin-etal-2010-panlex},\footnote{\url{https://panlex.org/}} to extend the language coverage of multilingual BERT (mBERT: \citet{devlin-etal-2019-bert}). They further pretrained mBERT using synthetic texts generated through word-to-word translation, resulting in improvements in named-entity recognition tasks. Drawing inspiration from their work, our study aims to reassess the utility of sentiment lexicons in sentiment analysis tasks, which were once a prominent feature in sentiment analysis prior to the advent of pre-trained language models. Specifically, we seek to answer the following questions: (1) \textit{To what extent do sentiment lexicons boost sentiment analysis using pretrained language models?}; and (2) \textit{Do multilingual sentiment lexicons improve the multilingual generalizability of sentiment analysis, particularly in low-resource languages?}

We chose sentiment classification as the focus of our study for two 
reasons. First, there is a wealth of sentiment classification datasets across diverse languages, allowing us to conduct experiments across 34 languages, 
including 6 high/medium-resource languages, 25 low-resource languages, and 3 code-switching language pairs. Secondly, compared to other semantic tasks such as hate speech detection \cite{schmidt-wiegand-2017-survey,rottger-etal-2021-hatecheck} and emotion recognition \cite{abdul-mageed-ungar-2017-emonet,sosea-caragea-2020-canceremo}, sentiment lexicons have been  studied extensively and are well-established in the field.

Cross-lingual transfer in sentiment classification is a prime case of low-resource NLP. However, existing research has predominantly focused on high/medium-resource languages 
 \citep{gupta-etal-2021-task,fei-li-2020-cross,lampleunsupervised}, relying on sentence-level sentiment datasets in English. In this paper, we showcase how models trained on English datasets are suboptimal for low-resource languages, and introduce lexicon-based pretraining that improves multilingual sentiment modeling. Our contributions can be summarized as follows:
\begin{itemize}
    \item 
    Our approach is arguably cost-effective since it relies exclusively on sentiment lexicons, 
    reducing the need for sentence-level sentiment annotation in any language, and sentence-level machine or human translation for low-resource languages, which can be challenging to access.
    \item We continue model pretraining using sentiment lexicons across 109 languages (see Figure~\ref{fig:overview}), and demonstrate strong zero-shot performance in low-resource languages, particularly in low-resource languages that are not covered by the multilingual lexicons, and in code-mixing texts that include high-resource languages. Our approach outperforms English models fine-tuned on sentence-level sentiment datasets, as well as large language models such as XGLM \cite{lin2021few}, BLOOMZ \cite{muennighoff2022crosslingual}, and GPT--3.5 \cite{ouyang2022training}.
    \item We conduct
    comprehensive experiments in two sentiment classification scenarios: binary 
    and 3-way classification. For each scenario, we benchmark two pretraining strategies: regression and classification. Unlike regression, the classification-based approach eases the constraint of determining the neutral class boundary before performing inference in 3-way classification in zero-shot setting.
    
    

\end{itemize}

\section{Related Work}
\label{sec:related}

We briefly review three subtopics that are pertinent to this work: (1) sentiment lexicons, (2) cross-lingual adaptation for sentiment analysis, and (3) sentiment analysis in low-resource languages.

\paragraph{Sentiment Lexicons} A sentiment lexicon is a curated collection of words and phrases that are classified as bearing positive or negative polarity. Such lexicons have applications in fields including NLP, cognitive science, psychology, and social science \cite{kiritchenko2014sentiment,mohammad-2018-obtaining}. There are two broad approaches to creating a sentiment lexicon: (1) direct annotation \cite{nielsen2011new,baccianella-etal-2010-sentiwordnet} --- have annotators assign sentiment scores to individual words on a rating scale, typically ranging from $-$5 (indicating very negative) to $+$5 (indicating very positive), or based on positive/negative categorical labels \cite{liu2005opinion}; and (2) best--worst scaling (BWS: \citet{kiritchenko2014sentiment,kiritchenko-etal-2016-semeval,mohammad-2018-obtaining}) --- have annotators select the most positive and least positive word from a collection of $n$ words, and infer a sentiment score based on the global rank of all words in the collection. BWS is considered more reliable than direct annotation as it helps mitigate annotator bias when assigning sentiment scores to individual words.

These sentiment lexicons have been constructed predominantly for English, but they also exist for languages such as Indonesian \cite{koto2017inset}, Arabic \cite{kiritchenko-etal-2016-semeval}, Persian \cite{dashtipour2016persent}, Dutch \cite{moors2013norms}, and Spanish \cite{redondo2007spanish}. In this work, we use NRC-VAD \cite{mohammad-2018-obtaining}, which is the largest English lexicon (19,965 words) and was built using the BWS method. Existing non-English lexicons are not just limited in size, they were also generally curated using a less reliable method (i.e.,~direct annotation). We instead use multilingual NRC-VAD lexicons in 108 languages, which is created by the original authors of NRC-VAD via Google Translate.\footnote{The approach aligns with the utilization of bilingual lexicons such as Panlex, as demonstrated in \citet{wang-etal-2022-expanding}.}


\paragraph{Cross-lingual Transfer in Sentiment Analysis} Most previous studies have primarily focused on cross-lingual adaptation in sentiment analysis by transferring models trained on English sentences $x_i$ and sentiment labels $y_i$ to other languages. \citet{abdalla-hirst-2017-cross} developed 
a mapper function to convert non-English word2vec embeddings to the English embedding space \cite{Mikolov2013EfficientEO}. \citet{zhou-etal-2016-cross,zhou-etal-2016-attention-based,wan-2009-co,lambert-2015-aspect} translated English datasets to several languages, such as Chinese and Spanish, and performed joint training to improve the multilingual capabilities of the model. \citet{fei-li-2020-cross} combined sentiment networks with unsupervised machine translation \cite{lampleunsupervised}, and 
\citet{meng-etal-2012-cross,jain-batra-2015-cross} have used unlabeled parallel texts in two languages to learn multilingual sentiment embeddings. 
In more recent work, \citet{sun-etal-2021-cross} used linguistic features such as language context, figurative language, and the lexification of emotional concepts to improve cross-lingual transfer, while \citet{zhang-etal-2021-cross} introduced a representation transformation technique from source to target languages which requires labeled English and non-English datasets. 

Cross-lingual transfer in previous work relies on sentence-level labeled English datasets, and has been evaluated on high/medium-resource languages. In this work, we do not use sentence-level labeled datasets, but solely lexicons, and test our methods on low-resource languages. To the best of our knowledge, our work constitutes the first effort to perform massively multilingual sentiment pretraining using lexicons.





\paragraph{Sentiment Analysis in Low-resource Languages} Most work in sentiment analysis has been applied to high/medium-resource languages, such as English \cite{nielsen2011new,baccianella-etal-2010-sentiwordnet,koto2015comparative}, Chinese \cite{zhou-etal-2016-cross,zhou-etal-2016-attention-based}, Japanese \cite{bataa-wu-2019-investigation}, and Indonesian \cite{koto2017inset,koto-etal-2021-indobertweet}. There also exists a small body 
of work on sentiment analysis for low-resource languages. First, NusaX \cite{winata-etal-2023-nusax} is a parallel sentiment analysis dataset that comprises 10 local Indonesian languages, along with
Indonesian and English translations.  SemEval-2023 \cite{muhammad-etal-2022-naijasenti,muhammad2023semeval} released sentiment analysis datasets for 14 African languages.
In other work, \citet{sirajzade-etal-2020-annotation} annotated Luxembourgish sentences with sentiment labels, and \citet{ali-etal-2021-creating} built a sentiment lexicon for Sindhi. In this study, we include the low-resource languages of NusaX and the 14 African languages from SemEval-2023 among our test sets.

\section{Pretraining with Sentiment Lexicons}

\subsection{Background and Problem Definition}

Prior research \cite[e.g.,][]{zhou-etal-2016-cross,zhang-etal-2021-cross} typically assumes access to sentence-level annotated data in a source language, often English, for zero-shot cross-lingual transfer to a target language. In this work, we define zero-shot as a setting where there is no sentence-level annotated data available in the source or target languages. Instead, we use the multilingual NRC-VAD lexicon \cite{mohammad-2018-obtaining} which comprises words $\{w_1, w_2, .., w_n\}$ manually annotated with valence $\{v_1, v_2, .., v_n\}$, arousal $\{a_1, a_2, .., a_n\}$, and dominance $\{d_1, d_2, .., d_n\}$ scores. In this work, we train only over the valence scores $v_i$, and normalize them from a range of $[0,1]$ to $[-5,5]$. 

Valence represents the degree of positiveness-negativeness/pleasure-displeasure and has been demonstrated to have a strong correlation with sentiment classification \cite{mohammad-2018-obtaining}. While the valence scores are suitable for regression, we also introduce valence classes $\{s_1, s_2, .., s_n\}$ that are derived from the valence score $v_i$. 
For 3-way classification we set the neutral class to $[-1,1)$, while we set 0 as the boundary between the positive and negative classes in the binary setting.

As illustrated in Figure~\ref{fig:overview}, we fine-tune  multilingual models \cite{devlin-etal-2019-bert,conneau-etal-2020-unsupervised,liu-etal-2020-multilingual-denoising,xue-etal-2021-mt5} on the parallel NRC-VAD lexicon in 109 languages. We specifically use average pooling over all tokens prior to the regression or classification layer. During zero-shot inference, we used fine-tuned models to predict sentiment labels at the sentence level.

\begin{table*}[!t]
    \centering
    \resizebox{\linewidth}{!}{
        \begin{tabular}{llcccccccccrr}
        \toprule
        \multicolumn{2}{c}{\multirow{2}{*}{\textbf{Lang}}} & \multicolumn{7}{c}{\textbf{Models}} & \textbf{NRC} & \multirow{2}{*}{\textbf{Panlex}} & \multicolumn{2}{c}{\textbf{train/dev/test}}  \\
        \cmidrule{3-9} \cmidrule{12-13}
         & & \textbf{mBERT} & \textbf{XLM-R} & \textbf{mBART} & \textbf{mT5} & \textbf{BLOOMZ} & \textbf{XGLM} & \textbf{GPT--3.5}  & \textbf{VAD} & & {\textbf{3-way}} & {\textbf{Binary
         }}  \\
         \midrule
        \multirow{6}{*}{\rotatebox{90}{\textbf{High/Medium}}} & en & \ok & \ok & \ok & \ok & \ok & \ok & \ok & \ok & \ok & 8544/1101/2210 & 6920/872/1821\\
        & ar & \ok & \ok & \ok & \ok & \ok & \ok & \ok & \ok & \ok & 3151/351/619 & 2162/251/428\\
        & es & \ok & \ok & \ok & \ok & \ok & \ok & \ok & \ok & \ok & 4802/2443/7264 & 3279/1650/5298\\ 
        & ru & \ok & \ok & \ok & \ok & \ok & \ok & \ok & \ok & \ok & 4113/726/4534 & 1205/209/1000\\
        & id & \ok & \ok & \ok & \ok & \ok & \ok & \ok & \ok & \ok & 3638/399/1011 & 3638/399/1011\\
        & ja & \ok & \ok & \ok & \ok & \ok & \ok & \ok & \ok & \ok & 3888/1112/553 & 2959/851/414\\
        \midrule
        \multirow{10}{*}{\rotatebox{90}{\textbf{Low/NusaX}}} & ace & \no & \no & \no & \no & \no & \no & \no & \no & \ok & 500/100/400 & 381/76/304\\
        & ban & \no & \no & \no & \no & \no & \no & \no & \no & \ok & 500/100/400 & 381/76/304 \\
        & bbc & \no & \no & \no & \no & \no & \no & \no & \no & \ok & 500/100/400 & 381/76/304 \\
        & bjn & \no & \no & \no & \no & \no & \no & \no & \no & \ok & 500/100/400 & 381/76/304 \\
        & bug & \no & \no & \no & \no & \no & \no & \no & \no & \ok & 500/100/400 & 381/76/304 \\
        & jv & \ok & \ok & \no & \ok & \no & \ok & \ok & \ok & \ok & 500/100/400 & 381/76/304 \\
        & mad & \no & \no & \no & \no & \no & \no & \no & \no & \no & 500/100/400 & 381/76/304 \\
        & min & \ok & \no & \no & \no & \no & \no & \ok & \no & \ok & 500/100/400 & 381/76/304 \\
        & nij & \no & \no & \no & \no & \no & \no & \no & \no & \ok & 500/100/400 & 381/76/304 \\
        & su & \ok & \ok & \no & \ok & \no & \ok & \ok & \ok & \ok & 500/100/400 & 381/76/304 \\
        \midrule
        \multirow{14}{*}{\rotatebox{90}{\textbf{Low/African}}} & am & \no & \ok & \no & \ok & \no & \ok & \ok & \ok & \ok & 5984/1497/1999 & 2880/721/1775 \\
        & dz & \no & \no & \no & \no & \no & \no & \no & \no & \ok & 1651/414/958 & 1309/328/804 \\
        & ha & \no & \ok & \no & \ok & \no & \ok & \ok & \ok & \ok & 14172/2677/5303 & 9260/1781/3514 \\
        & ig & \no & \no & \no & \ok & \ok & \ok & \ok & \ok & \ok & 10192/1841/3682 & 5684/1030/2061 \\
        & kr & \no & \no & \no & \no & \ok & \no & \ok & \ok & \ok & 8522/2090/4515 & 2045/512/633 \\
        & ma & \no & \no & \no & \no & \no & \no & \no & \no & \ok & 5583/1215/2961 & 3422/745/1994 \\
        & pcm & \no & \no & \no & \no & \no & \no & \no & \no & \ok & 5121/1281/4154 & 5049/1260/3723 \\
        & pt-MZ & \no & \no & \no & \no & \no & \no & \no & \no & \no & 3063/767/3662 & 1463/367/1283\\
        & sw & \ok & \ok & \no & \ok & \ok & \ok & \ok & \ok & \ok & 1810/453/748 & 738/185/304 \\
        & ts & \no & \no & \no & \no & \ok & \no & \ok & \no & \ok & 804/203/254 & 668/168/211\\
        & twi & \no & \no & \no & \no & \ok & \no & \no & \no & \ok & 3481/388/949 & 2959/330/803 \\
        & uo & \ok & \no & \no & \ok & \ok & \ok & \ok & \ok & \ok & 8522/2090/4515 & 5414/1327/2899\\
        & or & \no & \ok & \no & \no & \no & \ok & \ok & \no & \ok & 316/80/2096 & 218/53/1195\\
        & tg & \no & \no & \no & \no & \no & \ok & \no & \no & \ok & 318/80/2000 & 221/55/1613 \\
        & aeb & \no & \no & \no & \no & \no & \no & \no & \no & \ok & 4500/250/250 & 4284/232/235\\
        \midrule
        \multirow{3}{*}{\rotatebox{90}{\textbf{CW}}} & en-es & \ok & \ok & \ok & \ok & \ok & \ok & \ok & \ok & \ok & 2449/306/307 & 1405/162/182 \\
        & en-ml & \ok & \ok & \ok & \ok & \ok & \ok & \ok & \ok & \ok & 2856/358/335 & 2856/358/335 \\
        & en-ta & \ok & \ok & \ok & \ok & \ok & \ok & \ok & \ok & \ok & 3233/401/398 & 3233/401/398 \\

        \bottomrule
        \end{tabular}
    }
    \caption{Languages used in this paper. ``\cmark'' (green) and ``\xmark'' (red) mean that the language has and has not been seen by the models or language resources. CW indicates code-switching text. The language coverage for GPT-3.5 is derived from GPT-3 \cite{brown2020gpt3}.}
    \label{tab:language}
\end{table*}

\subsection{Extending the Lexicon}

As shown in Table~\ref{tab:simple}, the original NRC-VAD lexicon \cite{mohammad-2018-obtaining} comprises 19,965 English words, and has been extended to 108 languages by the original author resulting in 2.1M parallel words/phrases.\footnote{\url{https://saifmohammad.com/WebPages/nrc-vad.html}}

In  Table~\ref{tab:language}, we provide an overview of the languages and datasets used in this paper, categorized into: (1) high/medium-resource languages; (2) NusaX, covering local Indonesian languages (low resource); (3) African languages from SemEval 2023 (low resource); and (4) code-switching texts. The high/medium-resource languages and individual languages present in the code-switching texts are covered by all pretrained models and the NRC-VAD multilingual lexicon. However, for NusaX and the African languages, a considerable number of them are not covered.

Language coverage of the NRC-VAD multilingual lexicons remains limited in 109 languages. Therefore, we opt to extend the NRC-VAD lexicon using the Panlex lexicon, a ``panlingual'' lexicon containing translation edges between many languages.  As shown in Table~\ref{tab:language}, only \texttt{mad} and \texttt{pt-MZ} are not covered by Panlex. Specifically, we focus on 15 languages that are not covered by NRC-VAD, and project the sentiment scores from English. Given an English word and its valence score pair  $(w_i^{\text{en}}, v_i)$, we first obtain the translation of $w_i^{\text{en}}$ in language $L$. For each translation word $\{w_{i_1}^{L},w_{i_2}^{L},.., w_{i_m}^{L}\}$ we assign $v_i$ as the corresponding sentiment score. In total, we add 20K low-resource lexemes from 15 languages, as detailed in the Appendix (Table~\ref{tab:added-panlex}).

\subsection{Filtering Lexemes}

\begin{figure}[ht!]
    \centering
    \includegraphics[width=\linewidth]{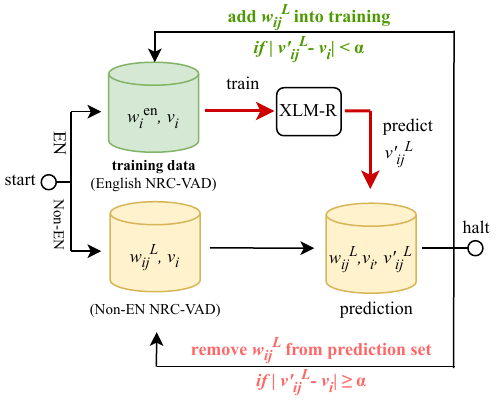} 
    \caption{Lexicon filtering pipeline.}
    \label{fig:filtering}
\end{figure}

Although translating lexemes is relatively easier and often more accurate than sentences, the sentiment score of the translated lexemes can be misleading because of word sense ambiguity. For example, the English word \ex{cottage} refers to a small house, while the Indonesian equivalent \ex{gubuk} \gl{shack} from the English-Indonesian lexicon may have more negative sentiment than \ex{cottage}.

To address the issue, we implement a filtering strategy illustrated in Figure~\ref{fig:filtering}. Initially, we train the English NRC-VAD $w_i^\text{en}$ with XLM-R \cite{conneau-etal-2020-unsupervised} using a regression approach. 
The training and validation data are split 80:20, with the model trained to predict the valence score $v_i$ based on the input word $w_i^\text{en}$. Subsequently, the model is used to predict the valence scores $v_i$ of additional lexemes (from the extended lexicons by \citet{mohammad-2018-obtaining} and Panlex). 
As a result, each word $w_{i_j}^{L}$ in the extended lexicon has two valence scores: the original score $v_i$ and the XLM score $\hat{v}_{i_j}^{L}$. All lexemes $w_{i_j}^{L}$ where the absolute difference $|\hat{v}_{i_j}^{L}-v_i|$ falls below a specified threshold $\alpha$ are added to the training and validation sets proportionally. This iterative process continues by training over the new extended lexicon until the number of additional words added to the training set becomes less than $\beta$.\footnote{We set the threshold $\alpha$ to 2.5, and $\beta$ to 1000.} 


\begin{table}[t]
    \centering
    \resizebox{0.9\linewidth}{!}{
        \begin{tabular}{llr}
        \toprule
        \textbf{id} & \textbf{Lexicon} & \textbf{Count} \\
        \midrule
        1 & Original NRC-VAD & 19,965 \\
        2 & (1) $+$ 108 translations & 2,176,185 \\
        3 & (2) $+$ Panlex extension (15 langs) & 2,196,252 \\
        4 & (3) $+$ Filtering & 2,071,691 \\
        \bottomrule
        \end{tabular}
    }
    \caption{Statistics of the original NRC-VAD lexicon, translations, Panlex extension, and filtering.}
    \label{tab:simple}
\end{table}

\section{Experiments}

\subsection{Data}
\label{sec:data}
As shown in Table~\ref{tab:language}, we use 34 languages in binary 
(positive, negative) and 3-way (positive, negative, neutral) classification scenarios. For binary 
classification, we simply remove sentences with neutral labels, resulting in a smaller dataset size. The 34 languages are grouped into 4 categories: 

\begin{itemize}
    \item \textbf{high/medium-resource languages}, including English (\texttt{en}: \citet{socher-etal-2013-recursive}), Arabic (\texttt{ar}: \citet{alturayeif-etal-2022-mawqif}), Spanish (\texttt{es}: \citet{garcia2020overview}), Russian (\texttt{ru}: \citet{loukachevitch2015sentirueval}), Indonesian (\texttt{id}: \citet{koto-etal-2020-indolem}), and Japanese (\texttt{ja}: \citet{hayashibe-2020-japanese}).
    \item \textbf{Low-resource languages from NusaX} \cite{winata-etal-2023-nusax}, consisting of 10 local Indonesian languages: Acehnese (\texttt{ace}), Balinese (\texttt{ban}), Batak Toba (\texttt{bbc}), Banjarase (\texttt{bjn}), Buginese (\texttt{bug}), Madurese (\texttt{mad}), Minangkabau (\texttt{min}), Javanese (\texttt{jv}), Ngaju (\texttt{nij}), and Sundanese (\texttt{su}).
    \item{ \textbf{Low-resource African languages}, based on the 14 languages of SemEval-2023 \cite{muhammad-etal-2022-naijasenti,muhammad2023semeval}: Amharic (\texttt{am}), Algerian Arabic (\texttt{dz}), Hausa (\texttt{ha}), Igbo (\texttt{ig}), Kinyarwanda (\texttt{kr}), Darija (\texttt{ma}), Nigerian Pidgin (\texttt{pcm}), Mozambique Portuguese (\texttt{pt-MZ}),\footnote{Mozambican Portuguese dialect differs in both pronunciation and colloquial vocabulary from standard European Portuguese.} Swahili (\texttt{sw}), Xitsonga (\texttt{ts}), Twi (\texttt{twi}), Yoruba (\texttt{yo}), Oromo (\texttt{or}), and Tigrinya (\texttt{tg}). We additionally include Tunisian Arabizi (\texttt{aeb}) from \citet{fourati-etal-2021-introducing}.}
    \item{\textbf{Code-switching texts}, involving English--Spanish (\texttt{en}--\texttt{es}: \citet{vilares-etal-2016-en}), English--Malayalam (\texttt{en}--\texttt{ml}: \citet{chakravarthi-etal-2020-sentiment}), and English--Tamil (\texttt{en}--\texttt{ta}: \citet{chakravarthi-etal-2020-corpus}})
\end{itemize}

\subsection{Set-Up}
\label{sec:setup}
 
We perform comprehensive evaluation over six multilingual encoder and encoder--decoder pretrained language models: (1)~mBERT$_\text{Base}$ \cite{devlin-etal-2019-bert}; (2)~XLM-R$_\text{Base}$ \cite{conneau-etal-2020-unsupervised}; (3)~XLM-R$_\text{Large}$ \cite{conneau-etal-2020-unsupervised}; (4)~mBART$_\text{Large}$ \cite{liu-etal-2020-multilingual-denoising}; (5)~mT5$_\text{Base}$ \cite{xue-etal-2021-mt5}; and (6)~mT5$_\text{Large}$ \cite{xue-etal-2021-mt5}. We evaluate different scenarios, including: (1) lexicon-based pretraining via regression vs.\ classification; and (2) binary 
vs.\ 3-way classification.

\paragraph{Lexicon-based pretraining} 
We conduct pretraining on the six multilingual pretrained language models using three combinations of multilingual lexicons: (1) NRC-VAD; (2) NRC-VAD $+$ Panlex; and (3) NRC-VAD $+$ Panlex $+$ filtering.
For regression-based pretraining, we use mean square error (MSE) loss, while for binary 
and 3-way classification, we use cross-entropy loss. Please see the Appendix for detailed hyper-parameter settings and computational resources.

\paragraph{Full Training, Few-shot and Zero-shot} Following the lexicon-based pretraining, we examine its impact on sentence-level sentiment analysis across three scenarios: (1) full training, (2) few-shot (training with limited data), and (3) zero-shot. For the first setting, we fine-tune the model with the complete training and development set of sentence-level sentiment data for each language listed in Table~\ref{tab:language}. For the second, we simulate few-shot training by randomly sampling 100 training and 50 development instances. To ensure robustness and account for variability, we repeat the experiment five times using different random seeds, and report the average performance. Please note that these first two settings are our preliminary experiments and we report the average scores of mBERT$_\text{Base}$ across 34 languages. Our main experiment in this work is zero-shot setting, simulating real-world scenarios for low-resource languages where no sentence-level sentiment data is available. For each of the six models, we present the average score for each language group in Section~\ref{sec:data}.


\paragraph{Baselines} In both full and few-shot training scenarios, the baseline consists of vanilla models without lexicon-based pretraining. For zero-shot setting, we compare our approaches with (1) models trained on SST datasets~\cite{socher-etal-2013-recursive} -- a sentence-level English sentiment data; and  (2) prompting via LLMs, including BLOOMZ (3B) \cite{muennighoff2022crosslingual}, XGLM (2.9B)~\cite{lin2021few}, and GPT--3.5 (175B) \cite{ouyang2022training}.\footnote{We do not include Llama-2~\cite{touvron2023llama} and Falcon~\cite{penedo2023refinedweb} as they are English-centric models.} The first baseline is zero-shot cross-lingual transfer, following prior work \cite{abdalla-hirst-2017-cross,zhang-etal-2021-cross} that used English as the main training language. For robustness, we fine-tuned the models with five different seeds for the first baseline. For the LLMs, we average the results of six English prompts as detailed in the Appendix. We report weighted macro-F1 scores for all experiments. 

Work discussed in Section~\ref{sec:related} is not suitable as a baseline due to the absence of word embeddings and machine translation systems in low-resource languages. For instance, \citet{abdalla-hirst-2017-cross} require 
word2vec embeddings in the target language, while \citet{zhou-etal-2016-cross, zhou-etal-2016-attention-based, wan-2009-co, lambert-2015-aspect, zhang-etal-2021-cross} rely on sentence-level machine translation. Additionally, \citet{meng-etal-2012-cross, jain-batra-2015-cross} require unlabeled parallel texts, which are not consistently available for low-resource languages.

\subsection{Results}

\begin{table}[t]
    \centering
    \resizebox{0.9\linewidth}{!}{
        \begin{tabular}{lcc}
        \toprule
        \textbf{Models} & \textbf{Binary
        } & \textbf{3-way} \\        
        \midrule
        \multicolumn{3}{l}{\textbf{Full training}} \\
        mBERT$_\text{Base}$ & 81.95 & 70.89 \\
        $+$ EN Lex.\ & 82.49 & 71.05 \\
        $+$ ML Lex.\ & 82.84 & 71.81 \\
        $+$ ML Lex.\ $+$ Panlex & \textbf{83.40} & 71.82 \\
        $+$ ML Lex.\ $+$ Panlex $+$ Filtering & \textbf{83.39} & \textbf{71.98} \\
        \midrule
        \multicolumn{3}{l}{\textbf{Training with limited data}} \\
        mBERT$_\text{Base}$ & 68.84 & 56.76 \\
        $+$ EN Lex.\ & 72.47 & 60.58 \\
        $+$ ML Lex.\ & 75.05 & 61.42 \\
        $+$ ML Lex.\ $+$ Panlex & 75.34 & 61.78 \\
        $+$ ML Lex.\ $+$ Panlex $+$ Filtering & \textbf{75.39} & \textbf{61.92} \\

        \bottomrule
        \end{tabular}
    }
    \caption{Preliminary results, based on averaged macro-F1 scores across 34 languages. ``EN Lex.''\ and ``ML Lex.''\ indicate the English and multilingual NRC-VAD lexicons.}
    \label{tab:early_result}
\end{table}

\begin{table*}[t]
    \centering
    \resizebox{\linewidth}{!}{
        \begin{tabular}{lccccccccccc}
        \toprule
            \multirow{2}{*}{\textbf{Model}} & \multicolumn{5}{c}{\textbf{Binary
            }} & & \multicolumn{5}{c}{\textbf{3-way}} \\

            \cmidrule{2-6} \cmidrule{8-12}
            & \textbf{HM-R} & \textbf{NusaX} & \textbf{African} & \textbf{CS} & \textbf{AVERAGE} &  & \textbf{HM-R} & \textbf{NusaX} & \textbf{African} & \textbf{CS} & \textbf{AVERAGE} \\
            \midrule
            XGLM (2.9B) & 59.66 & 49.34 & 42.50 & 52.61 & 51.03 &  & 38.09 & 33.47 & 25.72 & 50.08 & 36.84 \\
            BLOOMZ (3B) & \underline{77.82} & \underline{69.85} & \underline{54.92} & 45.89 & 62.12 &  & 48.43 & \underline{48.89} & 33.81 & 35.85 & 41.74 \\
            GPT--3.5 (175B) & 77.50 & 63.90 & 53.82 & \underline{73.66} & \underline{67.22} &  & \underline{67.65} & 48.50 & \underline{38.13} & \underline{50.41} & \underline{51.17} \\
            \midrule
            \textbf{mBERT$_\text{Base}$ (110M)} &  &  &  &  &  &  &  &  &  &  &  \\
            $+$ SST (sentence-level data) & 66.87 & 44.96 & 46.56 & 44.50 & 50.72 &  & 46.89 & 28.94 & 27.80 & 24.85 & 32.12 \\
            $+$ ML Lex.\ & 74.57 & \underline{67.92} & 57.79 & 69.43 & 67.43 &  & \underline{55.72} & \underline{44.18} & 35.08 & 60.14 & {48.78} \\
            $+$ ML Lex.\ $+$ Panlex & \underline{74.93} & 66.71 & \underline{58.99} & 71.58 & \underline{68.05} &  & 55.42 & 43.47 & \underline{35.13} & \underline{61.27} & \underline{48.82} \\
            $+$ ML Lex.\ $+$ Panlex $+$ Filtering & 74.74 & 63.95 & 58.00 & \underline{71.88} & 67.14 &  & 54.21 & 39.18 & 33.42 & 58.81 & 46.40 \\
            \midrule
            \textbf{XLM-R$_\text{Base}$ (270M)} &  &  &  &  &  &  &  &  &  &  &  \\
            $+$ SST (sentence-level data) & \underline{85.51} & 59.50 & 56.59 & 49.84 & 62.86 &  & 68.27 & 41.34 & 35.83 & 36.30 & 45.43 \\
            $+$ ML Lex.\ & 82.13 & 70.94 & 61.45 & 62.44 & 69.24 &  & 60.52 & 33.85 & 36.31 & 38.84 & 42.38 \\
            $+$ ML Lex.\ $+$ Panlex & 82.74 & \underline{73.59} & \underline{63.10} & 63.97 & 70.85 &  & 58.81 & \underline{46.34} & \underline{41.95} & \underline{60.31} & \underline{51.85} \\
            $+$ ML Lex.\ $+$ Panlex $+$ Filtering & 82.88 & 73.47 & 63.99 & \underline{72.50} & \underline{73.21} &  & \underline{61.28} & 33.33 & 34.87 & 40.13 & 42.40 \\
            \midrule
            \textbf{XLM-R$_\text{Large}$ (550M)} &  &  &  &  &  &  &  &  &  &  &  \\
            $+$ SST (sentence-level data) & \underline{\textbf{88.39}} & 73.00 & 61.75 & 54.38 & 69.38 &  & \underline{\textbf{70.57}} & 49.63 & 37.39 & 36.71 & 48.58 \\
            $+$ ML Lex.\ & 84.55 & \underline{78.01} & \underline{66.16} & 72.53 & \underline{75.31} &  & 61.78 & 45.95 & 41.84 & 63.52 & 53.27 \\
            $+$ ML Lex.\ $+$ Panlex & 84.89 & 70.79 & 63.85 & 76.47 & 74.00 &  & 64.38 & \underline{52.84} & 42.47 & \underline{64.73} & \underline{\textbf{56.10}} \\
            $+$ ML Lex.\ $+$ Panlex $+$ Filtering & 84.20 & 66.75 & 62.28 & \underline{\textbf{78.32}} & 72.89 &  & 64.74 & 46.67 & \underline{43.20} & 59.59 & 53.55 \\
            \midrule
            \textbf{mBART$_\text{Large}$ (600M)} &  &  &  &  &  &  &  &  &  &  &  \\
            $+$ SST (sentence-level data) & \underline{85.41} & 65.41 & 59.62 & 62.61 & 68.26 &  & \underline{66.58} & 31.80 & 27.99 & 31.52 & 39.47 \\
            $+$ ML Lex.\ & 83.26 & \underline{72.26} & \underline{62.89} & 74.10 & \underline{73.13} &  & 61.25 & \underline{41.88} & \underline{39.18} & \underline{54.43} & \underline{49.19} \\
            $+$ ML Lex.\ $+$ Panlex & 81.97 & 74.86 & 62.74 & 65.00 & 71.14 &  & 61.16 & 35.76 & 31.61 & 49.51 & 44.51 \\
            $+$ ML Lex.\ $+$ Panlex $+$ Filtering & 80.66 & 61.71 & 58.28 & \underline{78.02} & 69.67 &  & 57.48 & 30.50 & 30.24 & 40.77 & 39.75 \\
            \midrule
            \textbf{mT5$_\text{Base}$ (580M)} &  &  &  &  &  &  &  &  &  &  &  \\
            $+$ SST (sentence-level data) & \underline{83.16} & 55.33 & 57.18 & 48.39 & 61.02 &  & \underline{62.37} & 35.58 & 37.59 & 31.04 & 41.64 \\
            $+$ ML Lex.\ & 81.29 & \underline{75.84} & 67.45 & 73.63 & 74.55 &  & 59.27 & \underline{51.63} & 43.88 & \underline{60.04} & \underline{53.71} \\
            $+$ ML Lex.\ $+$ Panlex & 79.57 & 71.37 & 66.81 & 75.60 & 73.34 &  & 57.14 & 50.22 & \underline{44.62} & 58.51 & 52.62 \\
            $+$ ML Lex.\ $+$ Panlex $+$ Filtering & 82.24 & 75.52 & \underline{67.52} & \underline{76.33} & \underline{75.40} &  & 61.72 & 45.92 & 44.45 & 59.09 & 52.79 \\
            \midrule
            \textbf{mT5$_\text{Large}$} (1B) &  &  &  &  &  &  &  &  &  &  &  \\
            $+$ SST (sentence-level data) & \underline{84.74} & 60.68 & 58.67 & 47.91 & 63.00 &  & 48.05 & 31.67 & 31.53 & 24.75 & 34.00 \\
            $+$ ML Lex.\ & 83.69 & \underline{\textbf{78.26}} & 69.28 & 72.62 & 75.96 &  & \underline{62.15} & 51.84 & 44.42 & 61.43 & 54.96 \\
            $+$ ML Lex.\ $+$ Panlex & 82.78 & 76.70 & \underline{\textbf{70.05}} & 75.32 & \underline{\textbf{76.21}} &  & 59.59 & \underline{\textbf{53.96}} & \underline{\textbf{45.12}} & \underline{62.05} & \underline{55.18} \\
            $+$ ML Lex.\ $+$ Panlex $+$ Filtering & 81.35 & 73.37 & 68.04 & \underline{75.43} & 74.54 &  & 59.52 & 46.37 & 42.75 & 60.54 & 52.29 \\
        \bottomrule
        \end{tabular}
    }
    \caption{Full zero-shot results. The \underline{underlined} score indicates the highest performance within the respective group, while scores in \textbf{bold} indicate the best global performance. ``HM-R'' = high/medium-resource languages, excluding English, ``CS'' = code-switched text, and ``ML Lex.''\ indicates the multilingual NRC-VAD lexicon. ``SST (sentence-level data)'' is cross-lingual zero-shot transfer that is trained on English sentence-level sentiment data.}
    \label{tab:zeroshot_result}
\end{table*}

\paragraph{Preliminary Results: Full and Few-shot Training} Table~\ref{tab:early_result} shows the average performance of mBERT$_{\texttt{Base}}$ when training with full and limited training data at the sentence-level. Here we compare the vanilla multilingual model against four lexicon-based pretraining models, and its extensions (Panlex and Filtering). For each language in Table~\ref{tab:language}, we fine-tune the models and measure the macro-F1 score over the test set. For this preliminary experiment, we only use regression in lexicon-based pretraining for the binary 
and 3-way classification tasks. The results demonstrate that lexicon-based pretraining enhances performance, surpassing vanilla mBERT$_{\texttt{Base}}$ in both binary and 
3-way 
classification settings.
The proposed filtering method further slightly improves performance.

The improvements shown in Table~\ref{tab:early_result} are particularly noticeable in few-shot training, with increases of $+6.6$ and $+5.2$ for binary 
and 3-way classification, respectively. In the full training scenario, the increments are smaller, at only $+1.4$ and $+1.1$. These findings motivate us to further investigate zero-shot settings using all six multilingual models.

\paragraph{Zero-shot Results in High/Medium-Resource Languages} 

Table~\ref{tab:zeroshot_result} presents the averaged zero-shot performance of all models categorized by four language groups. The reported results use regression and classification in lexicon-based pretraining for binary 
and 3-way classification, respectively.
In the case of high/medium-resource languages (HM-R), English is excluded to ensure a fair comparison with models fine-tuned on the English SST dataset. Overall, we observe that multilingual models fine-tuned on SST tend to perform the best in high/medium-resource languages, with mBERT$_\text{Base}$ being the exception. It is not surprising to see these multilingual models outperforming the LLMs (the first three rows), as they are specifically fine-tuned on a sentence-level dataset.

Interestingly, we also observe that most of the lexicon-based pretrained models substantially outperform the LLMs. For instance, XLM-R$_\text{Large}$  outperforms GPT--3.5 and XGLM by $+7$ and $+24.9$ in binary 
classification. In 3-way classification, GPT--3.5 tends to perform better than lexicon-based pretraining, while XGLM and BLOOMZ tend to perform poorly. It's important to note that our models are significantly smaller in size, and BLOOMZ has been fine-tuned on multilingual sentiment analysis datasets, such as Amazon reviews \cite{muennighoff2022crosslingual}. 

\paragraph{Zero-shot Results in Low Resource Languages}
For low-resource languages, models fine-tuned on SST (i.e., sentence-level English dataset) underperform lexicon-based pretraining by a wide margin in both binary 
and 3-way classification settings. Notably, despite its significantly smaller size, lexicon-based pretraining to outperform larger models like BLOOMZ and GPT--3.5. Among the models, mT5$_\text{Large}$ achieves the best performance in NusaX and African languages for both classification scenarios, with disparities ranging from $+8.5$ to $+15$ and $+5$ to $+7$ when compared to LLMs. The impact of incorporating Panlex and/or filtering varies across models, with notable improvements observed for XLM-R$_\text{Base}$ and mT5$_\text{Large}$.

Expanding the multilingual lexicon with Panlex tends to improve the zero-shot capability for 3-way classification. This can be attributed to the fact that NusaX and African languages have a relatively small number of new lexemes (9.5K and 8.5K, respectively). Moreover, Panlex has English as the primary source language, making it inadequate to capture the diversity of languages in our experiments.

Although adding Panlex with the filtering method showed improvements in the preliminary experiment (
see the full training results in Table~\ref{tab:early_result}), it does not enhance the zero-shot performance in NusaX and African languages. To investigate this, 
we conducted a manual analysis of 100 randomly-selected samples from the 124K filtered lexemes. We compared the original sentiment scores of the corresponding English lexicon with the predicted scores generated by our filtering model. Upon back-translating the non-English words to English, we found that 63 of the original scores were either correct or better than the predicted scores, 25 predicted scores were better than the original scores, and 12 were incorrect for both. Additionally, we identified 75 unique languages among the 100 samples, indicating that our English-centric filtering might not be effective in improving low-resource languages.

\paragraph{Zero-shot Results in Code-switched Text} Extending the lexicon with Panlex and the filtering method yields the best performance for code-switched text, surpassing LLMs and models fine-tuned on SST. In binary 
classification, our method achieves an average F1-score that is $+24.1$ higher than the models fine-tuned on SST, while in 3-way classification, our method achieves F1-scores that are $+10$ to $+20$ higher than LLMs, even though the individual languages in our code-switched texts are high-resource (i.e.,~English, Spanish, Tamil, and Malayalam).

\section{Analysis}

\begin{table}[t]
    \centering
    \resizebox{\linewidth}{!}{
        \begin{tabular}{lccccc}
        \toprule
        \multirow{2}{*}{\textbf{Models + ML Lex.}} & \multicolumn{2}{c}{\textbf{Binary}} && \multicolumn{2}{c}{\textbf{3-way}} \\
        \cmidrule{2-3} \cmidrule{5-6}
        & \textbf{Reg.} &\textbf{ Class.} && \textbf{Reg.} & \textbf{Class.} \\
        \midrule
        mBERT$_\text{Base}$ & \textbf{67.43} & 66.29 & & 48.62 & \textbf{48.78} \\
        XLM-R$_\text{Base}$ & \textbf{69.24} & 64.37 & & \textbf{49.27} & 42.38 \\
        XLM-R$_\text{Large}$ & \textbf{75.31} & 74.54 & & \textbf{53.69} & 53.27 \\
        mBART$_\text{Large}$ & \textbf{73.13} & 71.23 & & 49.08 & \textbf{49.19} \\
        mT5$_\text{Base}$ & \textbf{74.55} & 68.27 & & 19.84 & \textbf{53.71} \\
        mT5$_\text{Large}$ & \textbf{75.96} & 72.21 & & 20.94 & \textbf{54.96} \\
        \bottomrule
        \end{tabular}
    }
    \caption{Regression vs.\ classification in lexicon-based pretraining for zero-shot sentiment analysis.}
    \label{tab:reg_vs_clas}
\end{table}

\begin{table}[t]
    \centering
    \resizebox{\linewidth}{!}{
        \begin{tabular}{lcccc}
        \toprule
            \textbf{Model} & \textbf{Stance} & \textbf{Hate Speech} & \textbf{Emotion} \\
            \midrule
            \textit{Binary classes} \\
            mBERT$_\text{Base}$ & 70.27 & 69.11 & 58.96 \\
            $+$ EN Lex.\ & \textbf{73.25} & \textbf{71.35} & \textbf{75.89} \\
            \midrule
            \textit{Original classes} \\
            mBERT$_\text{Base}$ & 52.55 & 56.36 & 18.44 \\
            $+$ EN Lex.\ & \textbf{53.04} & \textbf{57.59} & \textbf{23.93} \\
        \bottomrule
        \end{tabular}
    }
    \caption{Lexicon-based pretraining performance (macro-F1) over stance detection, hate speech detection, and emotion classification. The results are based on the limited training data scenario.}
    \label{tab:analysis_generalization}
\end{table}

\paragraph{Regression vs.\ classification in lexicon-based pretraining}

In Table~\ref{tab:reg_vs_clas} we present the average performance across the four language groups to compare the effectiveness of lexicon-based pretraining in regression and classification tasks for both binary 
and 3-way classification. Our findings indicate that regression performs better for 
binary classification, while classification leads to better results for 3-way classification. However, regression in 3-way classification presents a challenge when determining the neutral class boundary during inference. In the zero-shot setting, we lack specific data for hyper-parameter tuning, leading us to arbitrarily set the neutral class boundaries to $-$1 and $+$1. Although this setting works reasonably well for XLM-R, it yields poor performance for mT5. 
A manual analysis of mT5's predictions revealed that they tend to cluster around zero. 

\paragraph{Performance over unseen low-resource languages}

\begin{figure}[t]
    \centering
    \includegraphics[width=\linewidth]{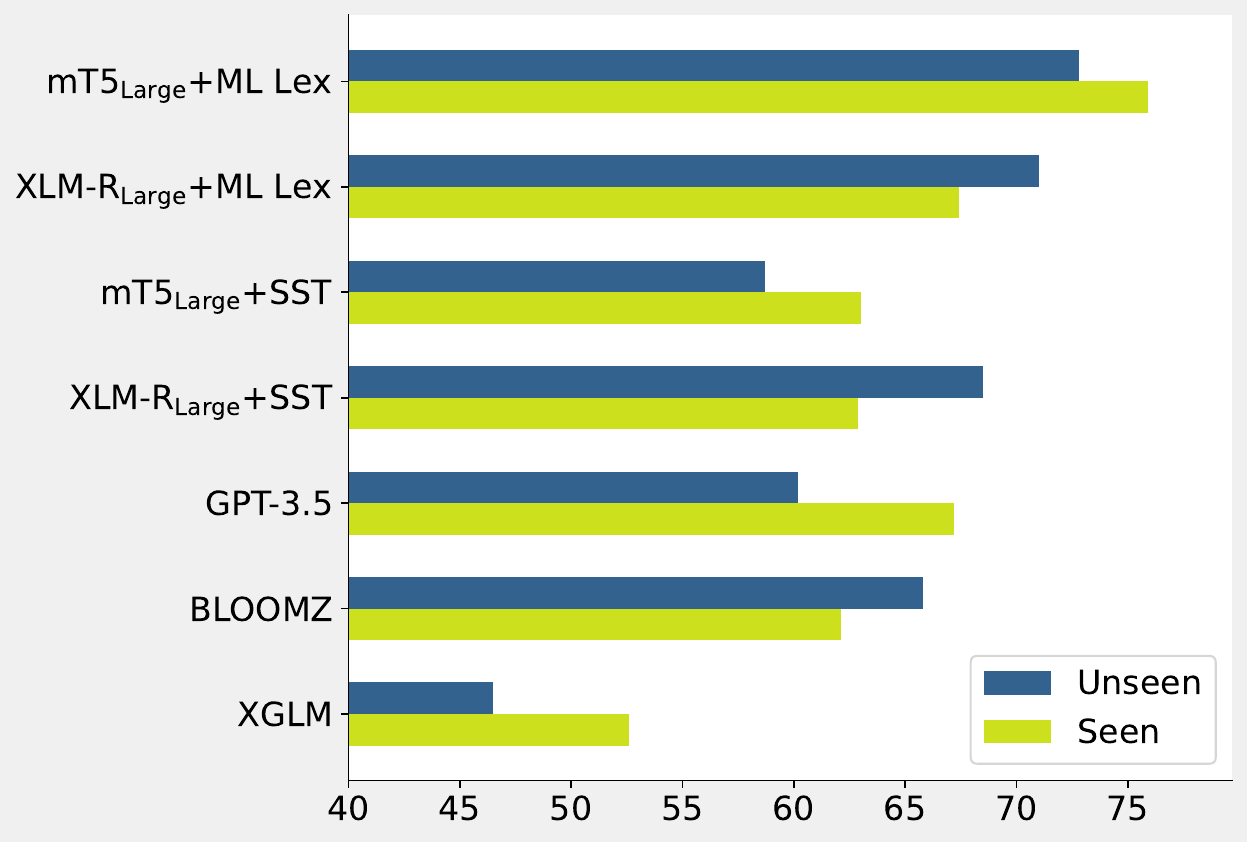} 
    \caption{Average zero-shot performance of seen and unseen languages in binary 
    classification across different models.}
    \label{fig:analysis}
\end{figure}

We compute the average results for languages that are completely unseen by all models, including 7 NusaX languages (\texttt{ace}, \texttt{ban}, \texttt{bbc}, \texttt{bjn}, \texttt{bug}, \texttt{mad}, \texttt{nij}) and 4 African languages (\texttt{dz}, \texttt{ma}, \texttt{pcm}, \texttt{pt-MZ}, \texttt{aeb}). We exclude lexicon-based pretraining with the Panlex extension since its performance is poor for low-resource languages. As a comparison, we include code-switched text as seen languages for the models. 
Figure~\ref{fig:analysis} presents the performance of XLM-R$_\text{Large}$ and mT5$_\text{Large}$, which outperform the LLMs and models fine-tuned on SST. This suggests that the multilingual sentiment lexicon is effective at enhancing language generalization for low-resource languages.

\section{Discussion}

Given the positive results, we explore the potential applicability of our methodology to NLP tasks beyond sentiment analysis, offering valuable directions for future research. We examine if the lexicon-based pretraining yields benefits in other semantic tasks, including stance detection \cite{li-etal-2021-p}, hate speech detection \cite{vidgen-etal-2021-learning}, and emotion classification \cite{demszky-etal-2020-goemotions}.

For each task, we take two datasets and perform experiments in few-shot training using mBERT$_\text{Base}$, following the setup described in Section~\ref{sec:setup}. Instead of using the multilingual lexicon, we use the English NRC-VAD lexicon since all the data is in English. For detailed information about the datasets and results, see the Appendix.
Table~\ref{tab:analysis_generalization} shows the average F1 scores on each task, demonstrating that lexicon-based pretraining boosts the performance of vanilla mBERT$_\text{Base}$. Particularly noteworthy are the substantial improvements in the emotion classification task, with increments of $+16.9$ and $+5.5$ for the binary and original class settings, respectively. These findings highlight the potential of sentiment lexicons for various semantic tasks, particularly in the context of investigating their effectiveness in low-resource languages in future works.

\section{Conclusion}
We have demonstrated the efficacy of employing a multilingual sentiment lexicon for achieving multilingual generalization in language model pretraining.  Without utilizing sentence-level datasets in any language, we provide compelling evidence of superior zero-shot performance in sentiment analysis tasks for low-resource languages, surpassing the performance of large language models. These findings open up new avenues for research in the realm of low-resource languages, not only for language understanding but also language generation tasks. Our results encourage further exploration and investigation of this exciting research direction.

\section*{Limitations}

This research focuses on general sentiment analysis, and we acknowledge that aspect-based sentiment analysis is a more fine-grained and expressive way of capturing sentiment, that warrants further exploration. Unfortunately, due to the scarcity of relevant datasets in low-resource languages, and task complexity, we were unable to explore aspect-based sentiment analysis in this work.

Regarding the proposed technique, we acknowledge four notable limitations. Firstly, due to the distinct nature of training (i.e.,~lexicon-level) and inference (i.e.,~sentence-level), our model may lack sensitivity to semantic complexity at the sentence level, encompassing nuances such as negation and sentences conveying multiple sentiments. One way to address this is to expand the NRC-VAD lexicon to include phrases, metaphors, culturally relevant words, and syntehtic sentences derived from the lexicon.  Secondly, our lexicon-based pretraining is solely based on valence scores, and there is an intriguing avenue to explore the inclusion of dominance and arousal scores. Thirdly, the use of machine translation systems for translating lexicons may introduce errors in both translation and sentiment scoring. While translating lexicons is arguably less complex than translating entire sentences, a comprehensive error analysis of the translated lexicons and Panlex words could offer valuable insights into the quality of the additional lexicons. Fourthly, our filtering method (Figure~\ref{fig:filtering}) proves less effective in certain scenarios due to its English-centric nature. This limitation arises because the initial filtering model is exclusively trained using English lexicons. To enhance this method, we propose that incorporating lexicons manually annotated in more diverse languages could significantly improve its efficacy.


\section*{Ethical Considerations}

When conducting sentiment analysis in low-resource languages, there are several important considerations that warrant reflection. First, it is crucial to ensure that the work benefits the local community rather than solely exploiting the language. In the era of large language models, the lack of computing resources often hinders the deployment of such systems in regions or countries where the language is spoken. Secondly, sentiment analysis can be subject to cultural ambiguity. Relying solely on European-centric multilingual models for sentiment prediction may introduce biases and produce inappropriate model predictions in certain cultural contexts. Therefore, cultural sensitivity and awareness are essential factors to address when conducting sentiment analysis in low-resource languages, which we leave for future work.


\bibliography{custom,anthology}

\begin{thebibliography}{69}
\expandafter\ifx\csname natexlab\endcsname\relax\def\natexlab#1{#1}\fi

\bibitem[{Abdalla and Hirst(2017)}]{abdalla-hirst-2017-cross}
Mohamed Abdalla and Graeme Hirst. 2017.
\newblock \href {https://aclanthology.org/I17-1051} {Cross-lingual sentiment
  analysis without (good) translation}.
\newblock In \emph{Proceedings of the Eighth International Joint Conference on
  Natural Language Processing (Volume 1: Long Papers)}, pages 506--515, Taipei,
  Taiwan. Asian Federation of Natural Language Processing.

\bibitem[{Abdul-Mageed and Ungar(2017)}]{abdul-mageed-ungar-2017-emonet}
Muhammad Abdul-Mageed and Lyle Ungar. 2017.
\newblock \href {https://doi.org/10.18653/v1/P17-1067} {{E}mo{N}et:
  Fine-grained emotion detection with gated recurrent neural networks}.
\newblock In \emph{Proceedings of the 55th Annual Meeting of the Association
  for Computational Linguistics (Volume 1: Long Papers)}, pages 718--728,
  Vancouver, Canada. Association for Computational Linguistics.

\bibitem[{Ali et~al.(2021)Ali, Ali, Dai, Kumar, Tumrani, and
  Xu}]{ali-etal-2021-creating}
Wazir Ali, Naveed Ali, Yong Dai, Jay Kumar, Saifullah Tumrani, and Zenglin Xu.
  2021.
\newblock \href {https://aclanthology.org/2021.wassa-1.20} {Creating and
  evaluating resources for sentiment analysis in the low-resource language:
  {S}indhi}.
\newblock In \emph{Proceedings of the Eleventh Workshop on Computational
  Approaches to Subjectivity, Sentiment and Social Media Analysis}, pages
  188--194, Online. Association for Computational Linguistics.

\bibitem[{Alturayeif et~al.(2022)Alturayeif, Luqman, and
  Ahmed}]{alturayeif-etal-2022-mawqif}
Nora~Saleh Alturayeif, Hamzah~Abdullah Luqman, and Moataz Aly~Kamaleldin Ahmed.
  2022.
\newblock \href {https://aclanthology.org/2022.wanlp-1.16} {Mawqif: A
  multi-label {A}rabic dataset for target-specific stance detection}.
\newblock In \emph{Proceedings of the The Seventh Arabic Natural Language
  Processing Workshop (WANLP)}, pages 174--184, Abu Dhabi, United Arab Emirates
  (Hybrid). Association for Computational Linguistics.

\bibitem[{Baccianella et~al.(2010)Baccianella, Esuli, and
  Sebastiani}]{baccianella-etal-2010-sentiwordnet}
Stefano Baccianella, Andrea Esuli, and Fabrizio Sebastiani. 2010.
\newblock \href
  {http://www.lrec-conf.org/proceedings/lrec2010/pdf/769_Paper.pdf}
  {{S}enti{W}ord{N}et 3.0: An enhanced lexical resource for sentiment analysis
  and opinion mining}.
\newblock In \emph{Proceedings of the Seventh International Conference on
  Language Resources and Evaluation ({LREC}'10)}, Valletta, Malta. European
  Language Resources Association (ELRA).

\bibitem[{Baldwin et~al.(2010)Baldwin, Pool, and
  Colowick}]{baldwin-etal-2010-panlex}
Timothy Baldwin, Jonathan Pool, and Susan Colowick. 2010.
\newblock \href {https://aclanthology.org/C10-3010} {{P}an{L}ex and {LEXTRACT}:
  Translating all words of all languages of the world}.
\newblock In \emph{Coling 2010: Demonstrations}, pages 37--40, Beijing, China.
  Coling 2010 Organizing Committee.

\bibitem[{Bataa and Wu(2019)}]{bataa-wu-2019-investigation}
Enkhbold Bataa and Joshua Wu. 2019.
\newblock \href {https://doi.org/10.18653/v1/P19-1458} {An investigation of
  transfer learning-based sentiment analysis in {J}apanese}.
\newblock In \emph{Proceedings of the 57th Annual Meeting of the Association
  for Computational Linguistics}, pages 4652--4657, Florence, Italy.
  Association for Computational Linguistics.

\bibitem[{Brown et~al.(2020)Brown, Mann, Ryder, Subbiah, Kaplan, Dhariwal,
  Neelakantan, Shyam, Sastry, Askell, Agarwal, Herbert-Voss, Krueger, Henighan,
  Child, Ramesh, Ziegler, Wu, Winter, Hesse, Chen, Sigler, Litwin, Gray, Chess,
  Clark, Berner, McCandlish, Radford, Sutskever, and Amodei}]{brown2020gpt3}
Tom Brown, Benjamin Mann, Nick Ryder, Melanie Subbiah, Jared~D Kaplan, Prafulla
  Dhariwal, Arvind Neelakantan, Pranav Shyam, Girish Sastry, Amanda Askell,
  Sandhini Agarwal, Ariel Herbert-Voss, Gretchen Krueger, Tom Henighan, Rewon
  Child, Aditya Ramesh, Daniel Ziegler, Jeffrey Wu, Clemens Winter, Chris
  Hesse, Mark Chen, Eric Sigler, Mateusz Litwin, Scott Gray, Benjamin Chess,
  Jack Clark, Christopher Berner, Sam McCandlish, Alec Radford, Ilya Sutskever,
  and Dario Amodei. 2020.
\newblock \href
  {https://proceedings.neurips.cc/paper_files/paper/2020/file/1457c0d6bfcb4967418bfb8ac142f64a-Paper.pdf}
  {Language models are few-shot learners}.
\newblock In \emph{Advances in Neural Information Processing Systems},
  volume~33, pages 1877--1901. Curran Associates, Inc.

\bibitem[{Chakravarthi et~al.(2020{\natexlab{a}})Chakravarthi, Jose,
  Suryawanshi, Sherly, and McCrae}]{chakravarthi-etal-2020-sentiment}
Bharathi~Raja Chakravarthi, Navya Jose, Shardul Suryawanshi, Elizabeth Sherly,
  and John~Philip McCrae. 2020{\natexlab{a}}.
\newblock \href {https://aclanthology.org/2020.sltu-1.25} {A sentiment analysis
  dataset for code-mixed {M}alayalam-{E}nglish}.
\newblock In \emph{Proceedings of the 1st Joint Workshop on Spoken Language
  Technologies for Under-resourced languages (SLTU) and Collaboration and
  Computing for Under-Resourced Languages (CCURL)}, pages 177--184, Marseille,
  France. European Language Resources association.

\bibitem[{Chakravarthi et~al.(2020{\natexlab{b}})Chakravarthi, Muralidaran,
  Priyadharshini, and McCrae}]{chakravarthi-etal-2020-corpus}
Bharathi~Raja Chakravarthi, Vigneshwaran Muralidaran, Ruba Priyadharshini, and
  John~Philip McCrae. 2020{\natexlab{b}}.
\newblock \href {https://aclanthology.org/2020.sltu-1.28} {Corpus creation for
  sentiment analysis in code-mixed {T}amil-{E}nglish text}.
\newblock In \emph{Proceedings of the 1st Joint Workshop on Spoken Language
  Technologies for Under-resourced languages (SLTU) and Collaboration and
  Computing for Under-Resourced Languages (CCURL)}, pages 202--210, Marseille,
  France. European Language Resources association.

\bibitem[{Chau et~al.(2020)Chau, Lin, and Smith}]{chau-etal-2020-parsing}
Ethan~C. Chau, Lucy~H. Lin, and Noah~A. Smith. 2020.
\newblock \href {https://doi.org/10.18653/v1/2020.findings-emnlp.118} {Parsing
  with multilingual {BERT}, a small corpus, and a small treebank}.
\newblock In \emph{Findings of the Association for Computational Linguistics:
  EMNLP 2020}, pages 1324--1334, Online. Association for Computational
  Linguistics.

\bibitem[{Conforti et~al.(2020)Conforti, Berndt, Pilehvar, Giannitsarou,
  Toxvaerd, and Collier}]{conforti-etal-2020-will}
Costanza Conforti, Jakob Berndt, Mohammad~Taher Pilehvar, Chryssi Giannitsarou,
  Flavio Toxvaerd, and Nigel Collier. 2020.
\newblock \href {https://doi.org/10.18653/v1/2020.acl-main.157}
  {Will-they-won{'}t-they: A very large dataset for stance detection on
  {T}witter}.
\newblock In \emph{Proceedings of the 58th Annual Meeting of the Association
  for Computational Linguistics}, pages 1715--1724, Online. Association for
  Computational Linguistics.

\bibitem[{Conneau et~al.(2020)Conneau, Khandelwal, Goyal, Chaudhary, Wenzek,
  Guzm{\'a}n, Grave, Ott, Zettlemoyer, and
  Stoyanov}]{conneau-etal-2020-unsupervised}
Alexis Conneau, Kartikay Khandelwal, Naman Goyal, Vishrav Chaudhary, Guillaume
  Wenzek, Francisco Guzm{\'a}n, Edouard Grave, Myle Ott, Luke Zettlemoyer, and
  Veselin Stoyanov. 2020.
\newblock \href {https://doi.org/10.18653/v1/2020.acl-main.747} {Unsupervised
  cross-lingual representation learning at scale}.
\newblock In \emph{Proceedings of the 58th Annual Meeting of the Association
  for Computational Linguistics}, pages 8440--8451, Online. Association for
  Computational Linguistics.

\bibitem[{Dashtipour et~al.(2016)Dashtipour, Hussain, Zhou, Gelbukh, Hawalah,
  and Cambria}]{dashtipour2016persent}
Kia Dashtipour, Amir Hussain, Qiang Zhou, Alexander Gelbukh, Ahmad~YA Hawalah,
  and Erik Cambria. 2016.
\newblock {PerSent}: A freely available {Persian} sentiment lexicon.
\newblock In \emph{Advances in Brain Inspired Cognitive Systems: 8th
  International Conference, BICS 2016, Beijing, China, November 28-30, 2016,
  Proceedings 8}, pages 310--320. Springer.

\bibitem[{Demszky et~al.(2020)Demszky, Movshovitz-Attias, Ko, Cowen, Nemade,
  and Ravi}]{demszky-etal-2020-goemotions}
Dorottya Demszky, Dana Movshovitz-Attias, Jeongwoo Ko, Alan Cowen, Gaurav
  Nemade, and Sujith Ravi. 2020.
\newblock \href {https://doi.org/10.18653/v1/2020.acl-main.372}
  {{G}o{E}motions: A dataset of fine-grained emotions}.
\newblock In \emph{Proceedings of the 58th Annual Meeting of the Association
  for Computational Linguistics}, pages 4040--4054, Online. Association for
  Computational Linguistics.

\bibitem[{Devlin et~al.(2019)Devlin, Chang, Lee, and
  Toutanova}]{devlin-etal-2019-bert}
Jacob Devlin, Ming-Wei Chang, Kenton Lee, and Kristina Toutanova. 2019.
\newblock \href {https://doi.org/10.18653/v1/N19-1423} {{BERT}: Pre-training of
  deep bidirectional transformers for language understanding}.
\newblock In \emph{Proceedings of the 2019 Conference of the North {A}merican
  Chapter of the Association for Computational Linguistics: Human Language
  Technologies, Volume 1 (Long and Short Papers)}, pages 4171--4186,
  Minneapolis, Minnesota. Association for Computational Linguistics.

\bibitem[{Fei and Li(2020)}]{fei-li-2020-cross}
Hongliang Fei and Ping Li. 2020.
\newblock \href {https://doi.org/10.18653/v1/2020.acl-main.510} {Cross-lingual
  unsupervised sentiment classification with multi-view transfer learning}.
\newblock In \emph{Proceedings of the 58th Annual Meeting of the Association
  for Computational Linguistics}, pages 5759--5771, Online. Association for
  Computational Linguistics.

\bibitem[{Founta et~al.(2018)Founta, Djouvas, Chatzakou, Leontiadis, Blackburn,
  Stringhini, Vakali, Sirivianos, and Kourtellis}]{founta2018large}
Antigoni Founta, Constantinos Djouvas, Despoina Chatzakou, Ilias Leontiadis,
  Jeremy Blackburn, Gianluca Stringhini, Athena Vakali, Michael Sirivianos, and
  Nicolas Kourtellis. 2018.
\newblock Large scale crowdsourcing and characterization of {Twitter} abusive
  behavior.
\newblock In \emph{Proceedings of the International AAAI Conference on Web and
  Social Media}.

\bibitem[{Fourati et~al.(2021)Fourati, Haddad, Messaoudi, BenHajhmida, Ben
  Elhaj~Mabrouk, and Naski}]{fourati-etal-2021-introducing}
Chayma Fourati, Hatem Haddad, Abir Messaoudi, Moez BenHajhmida, Aymen Ben
  Elhaj~Mabrouk, and Malek Naski. 2021.
\newblock \href {https://aclanthology.org/2021.wanlp-1.25} {Introducing a large
  {T}unisian {A}rabizi dialectal dataset for sentiment analysis}.
\newblock In \emph{Proceedings of the Sixth Arabic Natural Language Processing
  Workshop}, pages 226--230, Kyiv, Ukraine (Virtual). Association for
  Computational Linguistics.

\bibitem[{Garc{\'\i}a-Vega et~al.(2020)Garc{\'\i}a-Vega, D{\'\i}az-Galiano,
  Garc{\'\i}a-Cumbreras, Del~Arco, Montejo-R{\'a}ez, Jim{\'e}nez-Zafra,
  Mart{\'\i}nez~C{\'a}mara, Aguilar, Cabezudo, Chiruzzo
  et~al.}]{garcia2020overview}
Manuel Garc{\'\i}a-Vega, MC~D{\'\i}az-Galiano, MA~Garc{\'\i}a-Cumbreras, FMP
  Del~Arco, A~Montejo-R{\'a}ez, SM~Jim{\'e}nez-Zafra,
  E~Mart{\'\i}nez~C{\'a}mara, CA~Aguilar, MAS Cabezudo, L~Chiruzzo, et~al.
  2020.
\newblock Overview of {TASS} 2020: Introducing emotion detection.
\newblock In \emph{Proceedings of the Iberian Languages Evaluation Forum
  (IberLEF 2020) Co-Located with 36th Conference of the Spanish Society for
  Natural Language Processing (SEPLN 2020), M{\'a}laga, Spain}, pages 163--170.

\bibitem[{Gupta et~al.(2021)Gupta, Rallabandi, and
  Black}]{gupta-etal-2021-task}
Akshat Gupta, Sai~Krishna Rallabandi, and Alan~W Black. 2021.
\newblock \href {https://aclanthology.org/2021.dravidianlangtech-1.9}
  {Task-specific pre-training and cross lingual transfer for sentiment analysis
  in {D}ravidian code-switched languages}.
\newblock In \emph{Proceedings of the First Workshop on Speech and Language
  Technologies for Dravidian Languages}, pages 73--79, Kyiv. Association for
  Computational Linguistics.

\bibitem[{Hayashibe(2020)}]{hayashibe-2020-japanese}
Yuta Hayashibe. 2020.
\newblock \href {https://aclanthology.org/2020.lrec-1.843} {{J}apanese
  realistic textual entailment corpus}.
\newblock In \emph{Proceedings of the Twelfth Language Resources and Evaluation
  Conference}, pages 6827--6834, Marseille, France. European Language Resources
  Association.

\bibitem[{Jain and Batra(2015)}]{jain-batra-2015-cross}
Sarthak Jain and Shashank Batra. 2015.
\newblock \href {https://doi.org/10.18653/v1/D15-1016} {Cross lingual sentiment
  analysis using modified {BRAE}}.
\newblock In \emph{Proceedings of the 2015 Conference on Empirical Methods in
  Natural Language Processing}, pages 159--168, Lisbon, Portugal. Association
  for Computational Linguistics.

\bibitem[{Kiritchenko et~al.(2016)Kiritchenko, Mohammad, and
  Salameh}]{kiritchenko-etal-2016-semeval}
Svetlana Kiritchenko, Saif Mohammad, and Mohammad Salameh. 2016.
\newblock \href {https://doi.org/10.18653/v1/S16-1004} {{S}em{E}val-2016 task
  7: Determining sentiment intensity of {E}nglish and {A}rabic phrases}.
\newblock In \emph{Proceedings of the 10th International Workshop on Semantic
  Evaluation ({S}em{E}val-2016)}, pages 42--51, San Diego, California.
  Association for Computational Linguistics.

\bibitem[{Kiritchenko et~al.(2014)Kiritchenko, Zhu, and
  Mohammad}]{kiritchenko2014sentiment}
Svetlana Kiritchenko, Xiaodan Zhu, and Saif~M Mohammad. 2014.
\newblock Sentiment analysis of short informal texts.
\newblock \emph{Journal of Artificial Intelligence Research}, 50:723--762.

\bibitem[{Koto and Adriani(2015)}]{koto2015comparative}
Fajri Koto and Mirna Adriani. 2015.
\newblock A comparative study on {Twitter} sentiment analysis: Which features
  are good?
\newblock In \emph{Natural Language Processing and Information Systems: 20th
  International Conference on Applications of Natural Language to Information
  Systems, NLDB 2015, Passau, Germany, June 17-19, 2015, Proceedings 20}, pages
  453--457. Springer.

\bibitem[{Koto and Koto(2020)}]{koto-koto-2020-towards}
Fajri Koto and Ikhwan Koto. 2020.
\newblock \href {https://aclanthology.org/2020.paclic-1.17} {Towards
  computational linguistics in {M}inangkabau language: Studies on sentiment
  analysis and machine translation}.
\newblock In \emph{Proceedings of the 34th Pacific Asia Conference on Language,
  Information and Computation}, pages 138--148, Hanoi, Vietnam. Association for
  Computational Linguistics.

\bibitem[{Koto et~al.(2021)Koto, Lau, and
  Baldwin}]{koto-etal-2021-indobertweet}
Fajri Koto, Jey~Han Lau, and Timothy Baldwin. 2021.
\newblock \href {https://doi.org/10.18653/v1/2021.emnlp-main.833}
  {{I}ndo{BERT}weet: A pretrained language model for {I}ndonesian {T}witter
  with effective domain-specific vocabulary initialization}.
\newblock In \emph{Proceedings of the 2021 Conference on Empirical Methods in
  Natural Language Processing}, pages 10660--10668, Online and Punta Cana,
  Dominican Republic. Association for Computational Linguistics.

\bibitem[{Koto et~al.(2020)Koto, Rahimi, Lau, and
  Baldwin}]{koto-etal-2020-indolem}
Fajri Koto, Afshin Rahimi, Jey~Han Lau, and Timothy Baldwin. 2020.
\newblock \href {https://doi.org/10.18653/v1/2020.coling-main.66} {{I}ndo{LEM}
  and {I}ndo{BERT}: A benchmark dataset and pre-trained language model for
  {I}ndonesian {NLP}}.
\newblock In \emph{Proceedings of the 28th International Conference on
  Computational Linguistics}, pages 757--770, Barcelona, Spain (Online).
  International Committee on Computational Linguistics.

\bibitem[{Koto and Rahmaningtyas(2017)}]{koto2017inset}
Fajri Koto and Gemala~Y Rahmaningtyas. 2017.
\newblock Inset lexicon: Evaluation of a word list for {Indonesian} sentiment
  analysis in microblogs.
\newblock In \emph{2017 International Conference on Asian Language Processing
  (IALP)}, pages 391--394. IEEE.

\bibitem[{Lambert(2015)}]{lambert-2015-aspect}
Patrik Lambert. 2015.
\newblock \href {https://doi.org/10.3115/v1/P15-2128} {Aspect-level
  cross-lingual sentiment classification with constrained {SMT}}.
\newblock In \emph{Proceedings of the 53rd Annual Meeting of the Association
  for Computational Linguistics and the 7th International Joint Conference on
  Natural Language Processing (Volume 2: Short Papers)}, pages 781--787,
  Beijing, China. Association for Computational Linguistics.

\bibitem[{Lample et~al.(2018)Lample, Conneau, Denoyer, and
  Ranzato}]{lampleunsupervised}
Guillaume Lample, Alexis Conneau, Ludovic Denoyer, and Marc'Aurelio Ranzato.
  2018.
\newblock Unsupervised machine translation using monolingual corpora only.
\newblock In \emph{International Conference on Learning Representations},
  Vancouver, Canada.

\bibitem[{Li et~al.(2021)Li, Sosea, Sawant, Nair, Inkpen, and
  Caragea}]{li-etal-2021-p}
Yingjie Li, Tiberiu Sosea, Aditya Sawant, Ajith~Jayaraman Nair, Diana Inkpen,
  and Cornelia Caragea. 2021.
\newblock \href {https://doi.org/10.18653/v1/2021.findings-acl.208}
  {{P}-stance: A large dataset for stance detection in political domain}.
\newblock In \emph{Findings of the Association for Computational Linguistics:
  ACL-IJCNLP 2021}, pages 2355--2365, Online. Association for Computational
  Linguistics.

\bibitem[{Lin et~al.(2021)Lin, Mihaylov, Artetxe, Wang, Chen, Simig, Ott,
  Goyal, Bhosale, Du et~al.}]{lin2021few}
Xi~Victoria Lin, Todor Mihaylov, Mikel Artetxe, Tianlu Wang, Shuohui Chen,
  Daniel Simig, Myle Ott, Naman Goyal, Shruti Bhosale, Jingfei Du, et~al. 2021.
\newblock Few-shot learning with multilingual language models.
\newblock \emph{arXiv preprint arXiv:2112.10668}.

\bibitem[{Liu et~al.(2005)Liu, Hu, and Cheng}]{liu2005opinion}
Bing Liu, Minqing Hu, and Junsheng Cheng. 2005.
\newblock Opinion observer: analyzing and comparing opinions on the web.
\newblock In \emph{Proceedings of the 14th international conference on World
  Wide Web}, pages 342--351.

\bibitem[{Liu et~al.(2020)Liu, Gu, Goyal, Li, Edunov, Ghazvininejad, Lewis, and
  Zettlemoyer}]{liu-etal-2020-multilingual-denoising}
Yinhan Liu, Jiatao Gu, Naman Goyal, Xian Li, Sergey Edunov, Marjan
  Ghazvininejad, Mike Lewis, and Luke Zettlemoyer. 2020.
\newblock \href {https://doi.org/10.1162/tacl_a_00343} {Multilingual denoising
  pre-training for neural machine translation}.
\newblock \emph{Transactions of the Association for Computational Linguistics},
  8:726--742.

\bibitem[{Loukachevitch et~al.(2015)Loukachevitch, Blinov, Kotelnikov,
  Rubtsova, Ivanov, and Tutubalina}]{loukachevitch2015sentirueval}
Natalia Loukachevitch, Pavel Blinov, Evgeny Kotelnikov, Yulia Rubtsova,
  Vladimir Ivanov, and Elena Tutubalina. 2015.
\newblock {SentiRuEval}: testing object-oriented sentiment analysis systems in
  {Russian}.
\newblock In \emph{Proceedings of International Conference Dialog}, volume~2,
  pages 3--13.

\bibitem[{Meng et~al.(2012)Meng, Wei, Liu, Zhou, Xu, and
  Wang}]{meng-etal-2012-cross}
Xinfan Meng, Furu Wei, Xiaohua Liu, Ming Zhou, Ge~Xu, and Houfeng Wang. 2012.
\newblock \href {https://aclanthology.org/P12-1060} {Cross-lingual mixture
  model for sentiment classification}.
\newblock In \emph{Proceedings of the 50th Annual Meeting of the Association
  for Computational Linguistics (Volume 1: Long Papers)}, pages 572--581, Jeju
  Island, Korea. Association for Computational Linguistics.

\bibitem[{Mikolov et~al.(2013)Mikolov, Chen, Corrado, and
  Dean}]{Mikolov2013EfficientEO}
Tomas Mikolov, Kai Chen, Gregory~S. Corrado, and Jeffrey Dean. 2013.
\newblock Efficient estimation of word representations in vector space.
\newblock In \emph{International Conference on Learning Representations}.

\bibitem[{Mohammad(2018)}]{mohammad-2018-obtaining}
Saif Mohammad. 2018.
\newblock \href {https://doi.org/10.18653/v1/P18-1017} {Obtaining reliable
  human ratings of valence, arousal, and dominance for 20,000 {E}nglish words}.
\newblock In \emph{Proceedings of the 56th Annual Meeting of the Association
  for Computational Linguistics (Volume 1: Long Papers)}, pages 174--184,
  Melbourne, Australia. Association for Computational Linguistics.

\bibitem[{Mohammad et~al.(2018)Mohammad, Bravo-Marquez, Salameh, and
  Kiritchenko}]{mohammad-etal-2018-semeval}
Saif Mohammad, Felipe Bravo-Marquez, Mohammad Salameh, and Svetlana
  Kiritchenko. 2018.
\newblock \href {https://doi.org/10.18653/v1/S18-1001} {{S}em{E}val-2018 task
  1: Affect in tweets}.
\newblock In \emph{Proceedings of the 12th International Workshop on Semantic
  Evaluation}, pages 1--17, New Orleans, Louisiana. Association for
  Computational Linguistics.

\bibitem[{Moors et~al.(2013)Moors, De~Houwer, Hermans, Wanmaker, Van~Schie,
  Van~Harmelen, De~Schryver, De~Winne, and Brysbaert}]{moors2013norms}
Agnes Moors, Jan De~Houwer, Dirk Hermans, Sabine Wanmaker, Kevin Van~Schie,
  Anne-Laura Van~Harmelen, Maarten De~Schryver, Jeffrey De~Winne, and Marc
  Brysbaert. 2013.
\newblock Norms of valence, arousal, dominance, and age of acquisition for
  4,300 dutch words.
\newblock \emph{Behavior research methods}, 45:169--177.

\bibitem[{Muennighoff et~al.(2022)Muennighoff, Wang, Sutawika, Roberts,
  Biderman, Scao, Bari, Shen, Yong, Schoelkopf
  et~al.}]{muennighoff2022crosslingual}
Niklas Muennighoff, Thomas Wang, Lintang Sutawika, Adam Roberts, Stella
  Biderman, Teven~Le Scao, M~Saiful Bari, Sheng Shen, Zheng-Xin Yong, Hailey
  Schoelkopf, et~al. 2022.
\newblock Crosslingual generalization through multitask finetuning.
\newblock \emph{arXiv preprint arXiv:2211.01786}.

\bibitem[{Muhammad et~al.(2023)Muhammad, Abdulmumin, Yimam, Adelani, Ahmad,
  Ousidhoum, Ayele, Mohammad, Beloucif, and Ruder}]{muhammad2023semeval}
Shamsuddeen~Hassan Muhammad, Idris Abdulmumin, Seid~Muhie Yimam, David~Ifeoluwa
  Adelani, Ibrahim~Sa'id Ahmad, Nedjma Ousidhoum, Abinew~Ali Ayele, Saif~M.
  Mohammad, Meriem Beloucif, and Sebastian Ruder. 2023.
\newblock \href {https://arxiv.org/pdf/2304.06845.pdf} {{SemEval-2023 Task 12:
  Sentiment Analysis for African Languages (AfriSenti-SemEval)}}.
\newblock In \emph{Proceedings of the 17th {{International Workshop}} on
  {{Semantic Evaluation}} ({{SemEval-2023}})}. {Association for Computational
  Linguistics}.

\bibitem[{Muhammad et~al.(2022)Muhammad, Adelani, Ruder, Ahmad, Abdulmumin,
  Bello, Choudhury, Emezue, Abdullahi, Aremu, Jorge, and
  Brazdil}]{muhammad-etal-2022-naijasenti}
Shamsuddeen~Hassan Muhammad, David~Ifeoluwa Adelani, Sebastian Ruder,
  Ibrahim~Sa{'}id Ahmad, Idris Abdulmumin, Bello~Shehu Bello, Monojit
  Choudhury, Chris~Chinenye Emezue, Saheed~Salahudeen Abdullahi, Anuoluwapo
  Aremu, Al{\'\i}pio Jorge, and Pavel Brazdil. 2022.
\newblock \href {https://aclanthology.org/2022.lrec-1.63} {{N}aija{S}enti: A
  {N}igerian {T}witter sentiment corpus for multilingual sentiment analysis}.
\newblock In \emph{Proceedings of the Thirteenth Language Resources and
  Evaluation Conference}, pages 590--602, Marseille, France. European Language
  Resources Association.

\bibitem[{Nielsen(2011)}]{nielsen2011new}
Finn~{\AA}rup Nielsen. 2011.
\newblock A new anew: Evaluation of a word list for sentiment analysis in
  microblogs.
\newblock In \emph{Workshop on'Making Sense of Microposts: Big things come in
  small packages}, pages 93--98.

\bibitem[{Ouyang et~al.(2022)Ouyang, Wu, Jiang, Almeida, Wainwright, Mishkin,
  Zhang, Agarwal, Slama, Ray et~al.}]{ouyang2022training}
Long Ouyang, Jeffrey Wu, Xu~Jiang, Diogo Almeida, Carroll Wainwright, Pamela
  Mishkin, Chong Zhang, Sandhini Agarwal, Katarina Slama, Alex Ray, et~al.
  2022.
\newblock Training language models to follow instructions with human feedback.
\newblock \emph{Advances in Neural Information Processing Systems},
  35:27730--27744.

\bibitem[{Penedo et~al.(2023)Penedo, Malartic, Hesslow, Cojocaru, Cappelli,
  Alobeidli, Pannier, Almazrouei, and Launay}]{penedo2023refinedweb}
Guilherme Penedo, Quentin Malartic, Daniel Hesslow, Ruxandra Cojocaru,
  Alessandro Cappelli, Hamza Alobeidli, Baptiste Pannier, Ebtesam Almazrouei,
  and Julien Launay. 2023.
\newblock The {RefinedWeb} dataset for {Falcon LLM}: outperforming curated
  corpora with web data, and web data only.
\newblock \emph{arXiv preprint arXiv:2306.01116}.

\bibitem[{Pfeiffer et~al.(2020)Pfeiffer, Vuli{\'c}, Gurevych, and
  Ruder}]{pfeiffer-etal-2020-mad}
Jonas Pfeiffer, Ivan Vuli{\'c}, Iryna Gurevych, and Sebastian Ruder. 2020.
\newblock \href {https://doi.org/10.18653/v1/2020.emnlp-main.617} {{MAD-X}:
  {A}n {A}dapter-{B}ased {F}ramework for {M}ulti-{T}ask {C}ross-{L}ingual
  {T}ransfer}.
\newblock In \emph{Proceedings of the 2020 Conference on Empirical Methods in
  Natural Language Processing (EMNLP)}, pages 7654--7673, Online. Association
  for Computational Linguistics.

\bibitem[{Redondo et~al.(2007)Redondo, Fraga, Padr{\'o}n, and
  Comesa{\~n}a}]{redondo2007spanish}
Jaime Redondo, Isabel Fraga, Isabel Padr{\'o}n, and Montserrat Comesa{\~n}a.
  2007.
\newblock The spanish adaptation of anew (affective norms for english words).
\newblock \emph{Behavior research methods}, 39(3):600--605.

\bibitem[{R{\"o}ttger et~al.(2021)R{\"o}ttger, Vidgen, Nguyen, Waseem,
  Margetts, and Pierrehumbert}]{rottger-etal-2021-hatecheck}
Paul R{\"o}ttger, Bertie Vidgen, Dong Nguyen, Zeerak Waseem, Helen Margetts,
  and Janet Pierrehumbert. 2021.
\newblock \href {https://doi.org/10.18653/v1/2021.acl-long.4} {{H}ate{C}heck:
  Functional tests for hate speech detection models}.
\newblock In \emph{Proceedings of the 59th Annual Meeting of the Association
  for Computational Linguistics and the 11th International Joint Conference on
  Natural Language Processing (Volume 1: Long Papers)}, pages 41--58, Online.
  Association for Computational Linguistics.

\bibitem[{Schmidt and Wiegand(2017)}]{schmidt-wiegand-2017-survey}
Anna Schmidt and Michael Wiegand. 2017.
\newblock \href {https://doi.org/10.18653/v1/W17-1101} {A survey on hate speech
  detection using natural language processing}.
\newblock In \emph{Proceedings of the Fifth International Workshop on Natural
  Language Processing for Social Media}, pages 1--10, Valencia, Spain.
  Association for Computational Linguistics.

\bibitem[{Sirajzade et~al.(2020)Sirajzade, Gierschek, and
  Schommer}]{sirajzade-etal-2020-annotation}
Joshgun Sirajzade, Daniela Gierschek, and Christoph Schommer. 2020.
\newblock \href {https://aclanthology.org/2020.sltu-1.24} {An annotation
  framework for {L}uxembourgish sentiment analysis}.
\newblock In \emph{Proceedings of the 1st Joint Workshop on Spoken Language
  Technologies for Under-resourced languages (SLTU) and Collaboration and
  Computing for Under-Resourced Languages (CCURL)}, pages 172--176, Marseille,
  France. European Language Resources association.

\bibitem[{Socher et~al.(2013)Socher, Perelygin, Wu, Chuang, Manning, Ng, and
  Potts}]{socher-etal-2013-recursive}
Richard Socher, Alex Perelygin, Jean Wu, Jason Chuang, Christopher~D. Manning,
  Andrew Ng, and Christopher Potts. 2013.
\newblock \href {https://aclanthology.org/D13-1170} {Recursive deep models for
  semantic compositionality over a sentiment treebank}.
\newblock In \emph{Proceedings of the 2013 Conference on Empirical Methods in
  Natural Language Processing}, pages 1631--1642, Seattle, Washington, USA.
  Association for Computational Linguistics.

\bibitem[{Sosea and Caragea(2020)}]{sosea-caragea-2020-canceremo}
Tiberiu Sosea and Cornelia Caragea. 2020.
\newblock \href {https://doi.org/10.18653/v1/2020.emnlp-main.715}
  {{C}ancer{E}mo: A dataset for fine-grained emotion detection}.
\newblock In \emph{Proceedings of the 2020 Conference on Empirical Methods in
  Natural Language Processing (EMNLP)}, pages 8892--8904, Online. Association
  for Computational Linguistics.

\bibitem[{Sun et~al.(2021)Sun, Ahn, Park, Tsvetkov, and
  Mortensen}]{sun-etal-2021-cross}
Jimin Sun, Hwijeen Ahn, Chan~Young Park, Yulia Tsvetkov, and David~R.
  Mortensen. 2021.
\newblock \href {https://doi.org/10.18653/v1/2021.eacl-main.204}
  {Cross-cultural similarity features for cross-lingual transfer learning of
  pragmatically motivated tasks}.
\newblock In \emph{Proceedings of the 16th Conference of the European Chapter
  of the Association for Computational Linguistics: Main Volume}, pages
  2403--2414, Online. Association for Computational Linguistics.

\bibitem[{Touvron et~al.(2023)Touvron, Martin, Stone, Albert, Almahairi,
  Babaei, Bashlykov, Batra, Bhargava, Bhosale et~al.}]{touvron2023llama}
Hugo Touvron, Louis Martin, Kevin Stone, Peter Albert, Amjad Almahairi, Yasmine
  Babaei, Nikolay Bashlykov, Soumya Batra, Prajjwal Bhargava, Shruti Bhosale,
  et~al. 2023.
\newblock Llama 2: Open foundation and fine-tuned chat models.
\newblock \emph{arXiv preprint arXiv:2307.09288}.

\bibitem[{Vidgen et~al.(2021)Vidgen, Thrush, Waseem, and
  Kiela}]{vidgen-etal-2021-learning}
Bertie Vidgen, Tristan Thrush, Zeerak Waseem, and Douwe Kiela. 2021.
\newblock \href {https://doi.org/10.18653/v1/2021.acl-long.132} {Learning from
  the worst: Dynamically generated datasets to improve online hate detection}.
\newblock In \emph{Proceedings of the 59th Annual Meeting of the Association
  for Computational Linguistics and the 11th International Joint Conference on
  Natural Language Processing (Volume 1: Long Papers)}, pages 1667--1682,
  Online. Association for Computational Linguistics.

\bibitem[{Vilares et~al.(2016)Vilares, Alonso, and
  G{\'o}mez-Rodr{\'\i}guez}]{vilares-etal-2016-en}
David Vilares, Miguel~A. Alonso, and Carlos G{\'o}mez-Rodr{\'\i}guez. 2016.
\newblock \href {https://aclanthology.org/L16-1655} {{EN}-{ES}-{CS}: An
  {E}nglish-{S}panish code-switching {T}witter corpus for multilingual
  sentiment analysis}.
\newblock In \emph{Proceedings of the Tenth International Conference on
  Language Resources and Evaluation ({LREC}'16)}, pages 4149--4153,
  Portoro{\v{z}}, Slovenia. European Language Resources Association (ELRA).

\bibitem[{Wan(2009)}]{wan-2009-co}
Xiaojun Wan. 2009.
\newblock \href {https://aclanthology.org/P09-1027} {Co-training for
  cross-lingual sentiment classification}.
\newblock In \emph{Proceedings of the Joint Conference of the 47th Annual
  Meeting of the {ACL} and the 4th International Joint Conference on Natural
  Language Processing of the {AFNLP}}, pages 235--243, Suntec, Singapore.
  Association for Computational Linguistics.

\bibitem[{Wang et~al.(2022)Wang, Ruder, and Neubig}]{wang-etal-2022-expanding}
Xinyi Wang, Sebastian Ruder, and Graham Neubig. 2022.
\newblock \href {https://doi.org/10.18653/v1/2022.acl-long.61} {Expanding
  pretrained models to thousands more languages via lexicon-based adaptation}.
\newblock In \emph{Proceedings of the 60th Annual Meeting of the Association
  for Computational Linguistics (Volume 1: Long Papers)}, pages 863--877,
  Dublin, Ireland. Association for Computational Linguistics.

\bibitem[{Wang et~al.(2020)Wang, K, Mayhew, and
  Roth}]{wang-etal-2020-extending}
Zihan Wang, Karthikeyan K, Stephen Mayhew, and Dan Roth. 2020.
\newblock \href {https://doi.org/10.18653/v1/2020.findings-emnlp.240}
  {Extending multilingual {BERT} to low-resource languages}.
\newblock In \emph{Findings of the Association for Computational Linguistics:
  EMNLP 2020}, pages 2649--2656, Online. Association for Computational
  Linguistics.

\bibitem[{Winata et~al.(2023)Winata, Aji, Cahyawijaya, Mahendra, Koto,
  Romadhony, Kurniawan, Moeljadi, Prasojo, Fung, Baldwin, Lau, Sennrich, and
  Ruder}]{winata-etal-2023-nusax}
Genta~Indra Winata, Alham~Fikri Aji, Samuel Cahyawijaya, Rahmad Mahendra, Fajri
  Koto, Ade Romadhony, Kemal Kurniawan, David Moeljadi, Radityo~Eko Prasojo,
  Pascale Fung, Timothy Baldwin, Jey~Han Lau, Rico Sennrich, and Sebastian
  Ruder. 2023.
\newblock \href {https://aclanthology.org/2023.eacl-main.57} {{N}usa{X}:
  Multilingual parallel sentiment dataset for 10 {I}ndonesian local languages}.
\newblock In \emph{Proceedings of the 17th Conference of the European Chapter
  of the Association for Computational Linguistics}, pages 815--834, Dubrovnik,
  Croatia. Association for Computational Linguistics.

\bibitem[{Wolf et~al.(2020)Wolf, Debut, Sanh, Chaumond, Delangue, Moi, Cistac,
  Rault, Louf, Funtowicz, Davison, Shleifer, von Platen, Ma, Jernite, Plu, Xu,
  Le~Scao, Gugger, Drame, Lhoest, and Rush}]{wolf-etal-2020-transformers}
Thomas Wolf, Lysandre Debut, Victor Sanh, Julien Chaumond, Clement Delangue,
  Anthony Moi, Pierric Cistac, Tim Rault, Remi Louf, Morgan Funtowicz, Joe
  Davison, Sam Shleifer, Patrick von Platen, Clara Ma, Yacine Jernite, Julien
  Plu, Canwen Xu, Teven Le~Scao, Sylvain Gugger, Mariama Drame, Quentin Lhoest,
  and Alexander Rush. 2020.
\newblock \href {https://doi.org/10.18653/v1/2020.emnlp-demos.6} {Transformers:
  State-of-the-art natural language processing}.
\newblock In \emph{Proceedings of the 2020 Conference on Empirical Methods in
  Natural Language Processing: System Demonstrations}, pages 38--45, Online.
  Association for Computational Linguistics.

\bibitem[{Xia et~al.(2021)Xia, Zheng, Mukherjee, Shokouhi, Neubig, and
  Awadallah}]{xia-etal-2021-metaxl}
Mengzhou Xia, Guoqing Zheng, Subhabrata Mukherjee, Milad Shokouhi, Graham
  Neubig, and Ahmed~Hassan Awadallah. 2021.
\newblock \href {https://doi.org/10.18653/v1/2021.naacl-main.42} {{M}eta{XL}:
  Meta representation transformation for low-resource cross-lingual learning}.
\newblock In \emph{Proceedings of the 2021 Conference of the North American
  Chapter of the Association for Computational Linguistics: Human Language
  Technologies}, pages 499--511, Online. Association for Computational
  Linguistics.

\bibitem[{Xue et~al.(2021)Xue, Constant, Roberts, Kale, Al-Rfou, Siddhant,
  Barua, and Raffel}]{xue-etal-2021-mt5}
Linting Xue, Noah Constant, Adam Roberts, Mihir Kale, Rami Al-Rfou, Aditya
  Siddhant, Aditya Barua, and Colin Raffel. 2021.
\newblock \href {https://doi.org/10.18653/v1/2021.naacl-main.41} {m{T}5: A
  massively multilingual pre-trained text-to-text transformer}.
\newblock In \emph{Proceedings of the 2021 Conference of the North American
  Chapter of the Association for Computational Linguistics: Human Language
  Technologies}, pages 483--498, Online. Association for Computational
  Linguistics.

\bibitem[{Zhang et~al.(2021)Zhang, He, Peng, Bing, and
  Lam}]{zhang-etal-2021-cross}
Wenxuan Zhang, Ruidan He, Haiyun Peng, Lidong Bing, and Wai Lam. 2021.
\newblock \href {https://doi.org/10.18653/v1/2021.emnlp-main.727}
  {Cross-lingual aspect-based sentiment analysis with aspect term
  code-switching}.
\newblock In \emph{Proceedings of the 2021 Conference on Empirical Methods in
  Natural Language Processing}, pages 9220--9230, Online and Punta Cana,
  Dominican Republic. Association for Computational Linguistics.

\bibitem[{Zhou et~al.(2016{\natexlab{a}})Zhou, Wan, and
  Xiao}]{zhou-etal-2016-attention-based}
Xinjie Zhou, Xiaojun Wan, and Jianguo Xiao. 2016{\natexlab{a}}.
\newblock \href {https://doi.org/10.18653/v1/D16-1024} {Attention-based {LSTM}
  network for cross-lingual sentiment classification}.
\newblock In \emph{Proceedings of the 2016 Conference on Empirical Methods in
  Natural Language Processing}, pages 247--256, Austin, Texas. Association for
  Computational Linguistics.

\bibitem[{Zhou et~al.(2016{\natexlab{b}})Zhou, Wan, and
  Xiao}]{zhou-etal-2016-cross}
Xinjie Zhou, Xiaojun Wan, and Jianguo Xiao. 2016{\natexlab{b}}.
\newblock \href {https://doi.org/10.18653/v1/P16-1133} {Cross-lingual sentiment
  classification with bilingual document representation learning}.
\newblock In \emph{Proceedings of the 54th Annual Meeting of the Association
  for Computational Linguistics (Volume 1: Long Papers)}, pages 1403--1412,
  Berlin, Germany. Association for Computational Linguistics.

\end{thebibliography}
\bibliographystyle{acl_natbib}

\newpage

\appendix

\section{Additional lexemes from Panlex}
\label{app:panlex}

\begin{table}[ht!]
    \centering
    \resizebox{0.4\linewidth}{!}{
        \begin{tabular}{lr}
        \toprule
        \textbf{Language} & \textbf{\#words} \\
        \midrule
		  \texttt{ace} & 548 \\
            \texttt{aeb} & 257 \\
            \texttt{arq} & 91 \\
            \texttt{ary} & 1702 \\
            \texttt{ban} & 1435 \\
            \texttt{bbc} & 857 \\
            \texttt{bjn} & 377 \\
            \texttt{bug} & 1001 \\
            \texttt{gaz} & 1253 \\
            \texttt{min} & 5755 \\
            \texttt{nij} & 1326 \\
            \texttt{pcm} & 58 \\
            \texttt{tir} & 5367 \\
            \texttt{tso} & 27 \\
            \texttt{twi} & 13 \\
            \midrule
            \textbf{Total} & 20067 \\
        \bottomrule
        \end{tabular}
    }
    \caption{Total lexemes added from Panlex. For \texttt{min} we additionally extend the lexicon with a bilingual \texttt{min--id} lexicon from \citet{koto-koto-2020-towards}.}
    \label{tab:added-panlex}
\end{table}

\section{Languages in Multilingual Sentiment Lexicon}

The NRC-VAD lexicon was initially developed in English and later translated into 108 languages by the original author using the Google Translate API \cite{mohammad-2018-obtaining}. The lexicon covers a total of 109 languages, including English, as follows:

Afrikaans (\texttt{af}), Albanian (\texttt{sq}), Amharic (\texttt{am}), Arabic (\texttt{ar}), Armenian (\texttt{hy}), Azerbaijani (\texttt{az}), Basque (\texttt{eu}), Belarusian (\texttt{be}), Bengali (\texttt{bn}), Bosnian (\texttt{bs}), Bulgarian (\texttt{bg}), Catalan (\texttt{ca}), Cebuano (\texttt{ceb}), Chichewa (\texttt{ny}), Chinese-Simplified (\texttt{zh}), Chinese-Traditional (\texttt{zh}), Corsican (\texttt{co}), Croatian (\texttt{hr}), Czech (\texttt{cs}), Danish (\texttt{da}), Dutch (\texttt{nl}), English (\texttt{en}), Esperanto (\texttt{eo}), Estonian (\texttt{et}), Filipino (\texttt{fil}), Finnish (\texttt{fi}), French (\texttt{fr}), Frisian (\texttt{fy}), Galician (\texttt{gl}), Georgian (\texttt{ka}), German (\texttt{de}), Greek (\texttt{el}), Gujarati (\texttt{gu}), Haitian-Creole (\texttt{ht}), Hausa (\texttt{ha}), Hawaiian (\texttt{haw}), Hebrew (\texttt{he}), Hindi (\texttt{hi}), Hmong (\texttt{hmn}), Hungarian (\texttt{hu}), Icelandic (\texttt{is}), Igbo (\texttt{ig}), Indonesian (\texttt{id}), Irish (\texttt{ga}), Italian (\texttt{it}), Japanese (\texttt{ja}), Javanese (\texttt{jv}), Kannada (\texttt{kn}), Kazakh (\texttt{kk}), Khmer (\texttt{km}), Kinyarwanda (\texttt{rw}), Korean (\texttt{ko}), Kurdish-Kurmanji (\texttt{ku}), Kyrgyz (\texttt{ky}), Lao (\texttt{lo}), Latin (\texttt{la}), Latvian (\texttt{lv}), Lithuanian (\texttt{lt}), Luxembourgish (\texttt{lb}), Macedonian (\texttt{mk}), Malagasy (\texttt{mg}), Malay (\texttt{ms}), Malayalam (\texttt{ml}), Maltese (\texttt{mt}), Maori (\texttt{mi}), Marathi (\texttt{mr}), Mongolian (\texttt{mn}),
Myanmar-Burmese (\texttt{my}), Nepali (\texttt{ne}), Norwegian (\texttt{no}), Odia (\texttt{or}), Pashto (\texttt{ps}), Persian (\texttt{fa}), Polish (\texttt{pl}), Portuguese (\texttt{pt}), Punjabi (\texttt{pa}), Romanian (\texttt{ro}), Russian (\texttt{ru}), Samoan (\texttt{sm}), Scots-Gaelic (\texttt{sco}), Serbian (\texttt{sr}), Sesotho (\texttt{st}), Shona (\texttt{sn}),
Sindhi (\texttt{sd}), Sinhala (\texttt{si}), Slovak (\texttt{sk}), Slovenian (\texttt{sl}), Somali (\texttt{so}), Spanish (\texttt{es}), Sundanese (\texttt{su}), Swahili (\texttt{sw}), Swedish (\texttt{sv}), Tajik (\texttt{tg}), Tamil (\texttt{ta}), Tatar (\texttt{tt}),
Telugu (\texttt{te}), Thai (\texttt{th}), Turkish (\texttt{tr}), Turkmen (\texttt{tk}), Ukrainian (\texttt{uk}), Urdu (\texttt{ur}), Uyghur (\texttt{ug}), Uzbek (\texttt{uz}), Vietnamese (\texttt{vi}), Welsh (\texttt{cy}), Xhosa (\texttt{xh}), Yiddish (\texttt{yi}), Yoruba (\texttt{yo}), Zulu (\texttt{zu})

\section{Model Artifacts}

\begin{table}[ht!]
    \centering
    \resizebox{\linewidth}{!}{
        \begin{tabular}{lr}
        \toprule
        \textbf{Models (\#parameters)} & \textbf{Source} \\
        \midrule
        mBERT$_\text{Base}$ (110M) & \texttt{bert-base-multilingual-cased} \\
        XLM-R$_\text{Base}$ (270M) & \texttt{xlm-roberta-base} \\
        XLM-R$_\text{Large}$ (550M) & \texttt{xlm-roberta-large} \\
        mBART$_\text{Large}$ (600M) & \texttt{mbart-large-50} \\
        mT5$_\text{Base}$ (580M) & \texttt{google/mt5-base} \\
        mT5$_\text{Large}$ (1B) & \texttt{google/mt5-large} \\
        XGLM (2.9B) & \texttt{facebook/xglm-2.9B} \\
        BLOOMZ (3B) & \texttt{bigscience/bloomz-3b} \\
        GPT--3.5 (175B) & \texttt{text-davinci-003} \\
        \bottomrule
        \end{tabular}
    }
    \caption{With the exception of GPT--3.5 \cite{ouyang2022training}, all models used in this study were sourced from Huggingface \cite{wolf-etal-2020-transformers}.}
    \label{tab:models}
\end{table}

\section{Prompts}

We use six different prompts in evaluating large language models:
\begin{itemize}
    \item \texttt{[INPUT] \newline What would be the sentiment of the text above? [LABELS]}.
    \item \texttt{What is the sentiment of this text? \newline Text: [INPUT] \newline Sentiment: [LABELS]}.
    \item \texttt{Text: [INPUT]\newline Please classify the sentiment of above text: [LABELS]}.
    \item \texttt{[INPUT]\newline What would be the sentiment of the text above? [OPTIONS]? [LABELS]}.
    \item \texttt{What is the sentiment of this text?\newline Text: [INPUT]\newline Answer with [OPTIONS]: [LABELS]}.
    \item \texttt{Text: [INPUT]\newline Please classify the sentiment of above text. Answer with [OPTIONS]: [LABELS]}.
\end{itemize}
where \texttt{[INPUT]} is the input text, \texttt{[OPTIONS]} list all sentiment labels, and \texttt{[LABELS]} represent a sentiment class. For instance, given the text \ex{I love you} in binary 
classification, for the first prompt, we compare two normalized log-likelihood of: 
\begin{itemize}
    \item \texttt{I love you \newline What would be the sentiment of the text above? positive}
    \item \texttt{I love you \newline What would be the sentiment of the text above? negative}
\end{itemize}

\section{Detailed Experimental Results}
\label{app:detailed}

See Table~\ref{tab:a}, Table~\ref{tab:b}, Table~\ref{tab:c} for full results of binary 
classification, and Table~\ref{tab:a2}, Table~\ref{tab:b2}, Table~\ref{tab:c2} for full results of 3-way classification.

\section{Hyperparameters and Training Configurations}
For lexicon-based pretraining, we utilize a single 32GB A100 GPU. We set the initial learning rate to 2e-5 and the maximum number of epochs to 100. A patience value of 5 is used for early stopping, and a dropout rate of 0.2 is applied. Additionally, we set the maximum token length to 10. Different batch sizes are employed for each model: mBERT$_\text{Base}$=4000, XLM-R$_\text{Base}$=4000, XLM-R$_\text{Large}$=1000, mBART$_\text{Large}$=1000, mT5$_\text{Base}$=500, and mT5$_\text{Large}$=500.

For fine-tuning the mBERT model in both the full and limited training data scenarios, we also configure the initial learning rate to 2e-5 and set the maximum number of epochs to 20. A patience value of 5 is employed for early stopping, while a dropout rate of 0.2 is utilized. The maximum token length is set to 512, and a batch size of 32 is used. 
We also use these settings  when training the English SST baseline model for five different seeds.

\begin{table*}[ht]
    \centering
    \resizebox{0.9\linewidth}{!}{
        \begin{tabular}{lcccccccccc}
        \toprule
            \multirow{2}{*}{\textbf{Method}} & \multicolumn{6}{c}{\textbf{High/Medium-Resource}} & & \multicolumn{3}{c}{\textbf{Code-switching}} \\
            \cmidrule{2-7} \cmidrule{9-11}
             & \textbf{en} & \textbf{ar} & \textbf{es} & \textbf{ru} & \textbf{id} & \textbf{ja} &  & \textbf{en-es} & \textbf{en-ml} & \textbf{en-ta} \\
            \midrule
            \multicolumn{3}{l}{\textbf{Full training}} \\ 
            mBERT$_\text{Base}$ & 88.40 & 80.94 & 82.66 & 81.73 & 85.91 & 93.34 &  & 84.25 & 88.69 & 79.83 \\
            $+$ ML Lex.\ & 88.46 & 81.89 & 83.21 & 82.45 & 85.73 & 93.72 &  & 85.15 & 88.98 & 80.86 \\
            $+$ ML Lex.\ $+$ Panlex  & 88.69 & 82.15 & 83.59 & 84.87 & 86.06 & 94.58 &  & 87.77 & 89.58 & 79.73 \\
            $+$ ML Lex.\ $+$ Panlex $+$ Filtering & 88.53 & 81.46 & 83.24 & 85.95 & 85.92 & 94.07 &  & 86.78 & 89.28 & 81.11 \\
            \midrule
            \multicolumn{3}{l}{\textbf{Training with limited data}} \\ 
            mBERT$_\text{Base}$ & 63.83 & 71.49 & 67.55 & 70.04 & 71.15 & 80.07 &  & 65.69 & 72.05 & 77.95 \\
            $+$ ML Lex.\ & 76.66 & 73.41 & 78.55 & 75.44 & 76.45 & 84.74 &  & 80.55 & 79.97 & 77.78 \\
            $+$ ML Lex.\ $+$ Panlex  & 75.97 & 75.69 & 79.69 & 75.11 & 78.21 & 83.07 &  & 79.92 & 77.83 & 78.03 \\
            $+$ ML Lex.\ $+$ Panlex $+$ Filtering & 76.27 & 75.19 & 79.95 & 76.91 & 77.15 & 84.39 &  & 78.19 & 79.34 & 77.98 \\
            \midrule
            \multicolumn{8}{l}{\textbf{Zero-shot (LLMs)}} \\  
            XGLM (2.9B) & 51.18  & 54.37  & 53.09  & 55.08  & 56.32  & 79.45  &  & 55.85  & 50.50  & 51.47 \\
            BLOOMZ (3B) & 95.05  & 81.08  & 81.99  & 71.58  & 78.45  & 75.99  &  & 71.83  & 27.89  & 37.95 \\
            GPT--3.5 (175B) & 84.52  & 72.14  & 72.53  & 80.29  & 72.68  & 89.85  &  & 79.83  & 68.50  & 72.66 \\
            \midrule
            \multicolumn{8}{l}{\textbf{Zero-shot (SST and Lexicon-based pretraining with regression)}} \\  
            mBERT$_\text{Base}$ (110M) &  &  &  &  &  &  &  &  &  & \\
            $+$ fine-tuned on SST & 88.42  & 69.92  & 74.39  & 65.64  & 67.24  & 57.17  &  & 61.38  & 29.87  & 42.26 \\
            $+$ ML Lex.\ & 71.42  & 73.06  & 76.89  & 81.84  & 72.73  & 68.35  &  & 69.61  & 69.58  & 69.09 \\
            $+$ ML Lex.\ $+$ Panlex & 70.39  & 75.31  & 77.08  & 77.27  & 68.40  & 60.41  &  & 72.44  & 61.18  & 50.47 \\
            $+$ ML Lex.\ $+$ Panlex $+$ Filtering & 71.44  & 75.20  & 78.86  & 75.75  & 70.87  & 73.00  &  & 70.43  & 72.29  & 72.93 \\
            \hdashline
            XLM-R$_\text{Base}$ (270M) &  &  &  &  &  &  &  &  &  & \\
            $+$ fine-tuned on SST & 91.48  & 81.53  & 83.98  & 88.40  & 81.70  & 91.94  &  & 67.79  & 35.75  & 45.99 \\
            $+$ ML Lex.\ & 73.89  & 76.16  & 78.84  & 84.63  & 80.22  & 90.82  &  & 68.58  & 54.54  & 64.19 \\
            $+$ ML Lex.\ $+$ Panlex & 74.09  & 77.80  & 79.45  & 84.23  & 81.80  & 90.43  &  & 69.29  & 57.17  & 65.46 \\
            $+$ ML Lex.\ $+$ Panlex $+$ Filtering & 75.85  & 77.80  & 80.43  & 83.76  & 80.92  & 91.48  &  & 74.31  & 71.44  & 71.76 \\
            \hdashline
            XLM-R$_\text{Large}$ (550M) &  &  &  &  &  &  &  &  &  & \\
            $+$ fine-tuned on SST & 94.00  & 85.37  & 85.95  & 92.64  & 82.88  & 95.11  &  & 75.36  & 38.21  & 49.57 \\
            $+$ ML Lex.\ & 78.14  & 79.85  & 82.12  & 88.46  & 82.47  & 89.87  &  & 75.41  & 70.70  & 71.47 \\
            $+$ ML Lex.\ $+$ Panlex & 76.00  & 80.73  & 81.76  & 87.48  & 82.88  & 91.61  &  & 77.10  & 75.23  & 77.06 \\
            $+$ ML Lex.\ $+$ Panlex $+$ Filtering & 77.82  & 78.99  & 81.55  & 87.95  & 80.42  & 92.06  &  & 78.66  & 79.80  & 76.49 \\
            \hdashline
            mBART$_\text{Large}$ (600M) &  &  &  &  &  &  &  &  &  & \\
            $+$ fine-tuned on SST & 94.00  & 85.37  & 85.95  & 92.64  & 82.88  & 95.11  &  & 75.36  & 38.21  & 49.57 \\
            $+$ ML Lex.\ & 75.73  & 82.31  & 76.31  & 87.18  & 80.38  & 90.15  &  & 78.45  & 71.20  & 72.67 \\
            $+$ ML Lex.\ $+$ Panlex & 78.96  & 81.18  & 79.52  & 87.07  & 81.49  & 80.59  &  & 74.38  & 58.79  & 61.83 \\
            $+$ ML Lex.\ $+$ Panlex $+$ Filtering & 74.38  & 58.79  & 61.83  & 74.38  & 58.79  & 61.83  &  & 76.40  & 73.67  & 76.23 \\
            \hdashline
            mT5$_\text{Base}$ (580M) &  &  &  &  &  &  &  &  &  & \\
            $+$ fine-tuned on SST & 90.21  & 79.99  & 82.58  & 84.00  & 78.87  & 90.37  &  & 70.47  & 30.49  & 44.21 \\
            $+$ ML Lex.\ & 72.55  & 76.48  & 79.18  & 83.58  & 75.11  & 92.11  &  & 77.53  & 69.86  & 73.49 \\
            $+$ ML Lex.\ $+$ Panlex & 71.67  & 76.29  & 76.53  & 79.60  & 74.34  & 91.08  &  & 79.65  & 71.53  & 75.63 \\
            $+$ ML Lex.\ $+$ Panlex $+$ Filtering & 75.62  & 77.52  & 80.93  & 84.55  & 75.34  & 92.83  &  & 77.62  & 74.30  & 77.06 \\
            \hdashline
            mT5$_\text{Large}$ (1B) &  &  &  &  &  &  &  &  &  & \\
            $+$ fine-tuned on SST & 91.13  & 81.92  & 83.14  & 87.48  & 79.82  & 91.34  &  & 70.18  & 29.47  & 44.07 \\
            $+$ ML Lex.\ & 71.99  & 82.82  & 79.06  & 88.42  & 76.07  & 92.06  &  & 77.62  & 68.57  & 71.67 \\
            $+$ ML Lex.\ $+$ Panlex & 72.37  & 80.44  & 77.34  & 86.11  & 77.97  & 92.02  &  & 78.66  & 74.73  & 72.56 \\
            $+$ ML Lex.\ $+$ Panlex $+$ Filtering & 70.52  & 80.26  & 75.80  & 81.44  & 78.01  & 91.22  &  & 76.40  & 73.67  & 76.23 \\

        \bottomrule
        \end{tabular} 
    }
    \caption{All \textbf{binary 
    classification} results for high/medium-resource languages and code-switched texts.}
    \label{tab:a}
\end{table*}

\begin{table*}[ht]
    \centering
    \resizebox{\linewidth}{!}{
        \begin{tabular}{lcccccccccc}
        \toprule
            \multirow{2}{*}{\textbf{Method}} & \multicolumn{10}{c}{\textbf{NusaX}} \\
             \cmidrule{2-11}
             & \textbf{ace} & \textbf{ban} & \textbf{bbc} & \textbf{bjn} & \textbf{bug} & \textbf{jav} & \textbf{mad} & \textbf{min} & \textbf{nij} & \textbf{sun}\\
            \midrule
            \multicolumn{3}{l}{\textbf{Full training}} \\
            mBERT$_\text{Base}$ & 83.84  & 85.44  & 84.42  & 84.99  & 83.85  & 87.82  & 84.31  & 87.87  & 83.93  & 85.90 \\
            $+$ ML Lex.\ & 84.96  & 85.97  & 84.16  & 89.44  & 84.19  & 92.04  & 86.04  & 88.28  & 84.84  & 89.14 \\
            $+$ ML Lex.\ $+$ Panlex  & 85.32  & 87.26  & 85.58  & 88.65  & 83.64  & 92.56  & 85.51  & 88.88  & 86.82  & 89.79 \\
            $+$ ML Lex.\ $+$ Panlex $+$ Filtering & 84.33  & 87.22  & 84.52  & 89.83  & 83.73  & 93.35  & 85.17  & 89.33  & 87.66  & 90.06 \\
            \midrule
            \multicolumn{3}{l}{\textbf{Training with limited data}} \\  
            mBERT$_\text{Base}$ & 70.64  & 71.89  & 69.77  & 77.49  & 68.61  & 81.58  & 72.01  & 79.57  & 74.03  & 79.04 \\
            $+$ ML Lex.\ & 79.72  & 76.15  & 76.51  & 79.20  & 76.14  & 89.65  & 78.13  & 85.73  & 81.67  & 87.95 \\
            $+$ ML Lex.\ $+$ Panlex  & 80.70  & 78.22  & 76.84  & 83.21  & 75.60  & 91.36  & 80.86  & 85.31  & 81.00  & 86.55 \\
            $+$ ML Lex.\ $+$ Panlex $+$ Filtering & 78.85  & 77.13  & 76.57  & 83.70  & 75.57  & 90.37  & 79.22  & 87.27  & 82.88  & 85.84 \\
            \midrule
            \multicolumn{8}{l}{\textbf{Zero-shot (LLMs)}} \\ 
            XGLM (2.9B) & 48.19  & 53.01  & 41.04  & 53.20  & 39.45  & 55.61  & 51.38  & 53.45  & 46.49  & 51.55 \\
            BLOOMZ (3B) & 74.10  & 74.05  & 55.56  & 83.20  & 49.92  & 81.35  & 67.08  & 78.66  & 68.97  & 65.56 \\
            GPT--3.5 (175B) & 63.52  & 63.53  & 55.37  & 72.84  & 46.19  & 75.43  & 61.01  & 72.49  & 60.28  & 68.32 \\
            \midrule
            \multicolumn{8}{l}{\textbf{Zero-shot (SST and Lexicon-based pretraining with regression)}} \\  
            mBERT$_\text{Base}$ (110M) &  &  &  &  &  &  &  &  &  & \\
            $+$ fine-tuned on SST & 38.05  & 47.89  & 39.04  & 46.86  & 34.94  & 54.90  & 40.48  & 51.26  & 44.80  & 51.37 \\
            $+$ ML Lex.\ & 66.00  & 67.58  & 62.38  & 69.22  & 58.68  & 80.23  & 65.66  & 70.38  & 61.23  & 77.88 \\
            $+$ ML Lex.\ $+$ Panlex & 64.06  & 62.30  & 59.23  & 66.60  & 53.98  & 80.58  & 68.63  & 72.50  & 62.48  & 76.73 \\
            $+$ ML Lex.\ $+$ Panlex $+$ Filtering & 60.07  & 58.00  & 58.64  & 63.41  & 48.48  & 80.22  & 62.71  & 72.00  & 61.62  & 74.31 \\
            \hdashline
            XLM-R$_\text{Base}$ (270M) &  &  &  &  &  &  &  &  &  & \\
            $+$ fine-tuned on SST & 56.63  & 62.59  & 38.17  & 74.12  & 35.34  & 80.86  & 59.55  & 65.79  & 58.28  & 63.71 \\
            $+$ ML Lex.\ & 68.56  & 80.23  & 49.21  & 83.21  & 44.25  & 90.79  & 66.42  & 81.20  & 64.39  & 81.17 \\
            $+$ ML Lex.\ $+$ Panlex & 66.57  & 80.77  & 57.25  & 83.87  & 50.90  & 89.80  & 69.23  & 84.21  & 71.13  & 82.17 \\
            $+$ ML Lex.\ $+$ Panlex $+$ Filtering & 62.47  & 77.40  & 64.54  & 79.22  & 59.19  & 86.18  & 72.37  & 83.19  & 70.57  & 79.59 \\
            \hdashline
            XLM-R$_\text{Large}$ (550M) &  &  &  &  &  &  &  &  &  & \\
            $+$ fine-tuned on SST & 68.43  & 78.26  & 49.79  & 86.87  & 44.73  & 91.82  & 75.23  & 80.73  & 69.62  & 84.48 \\
            $+$ ML Lex.\ & 73.19  & 77.61  & 66.00  & 82.35  & 63.41  & 92.43  & 74.31  & 84.81  & 74.88  & 91.11 \\
            $+$ ML Lex.\ $+$ Panlex & 60.81  & 64.60  & 59.76  & 77.01  & 55.83  & 89.40  & 69.28  & 81.87  & 64.28  & 85.01 \\
            $+$ ML Lex.\ $+$ Panlex $+$ Filtering & 52.42  & 65.90  & 53.43  & 72.08  & 40.84  & 87.71  & 65.75  & 78.81  & 65.27  & 85.33 \\
            \hdashline
            mBART$_\text{Large}$ (600M) &  &  &  &  &  &  &  &  &  & \\
            $+$ fine-tuned on SST & 62.93  & 70.97  & 46.23  & 75.61  & 46.88  & 81.68  & 71.92  & 72.80  & 63.00  & 62.09 \\
            $+$ ML Lex.\ & 67.10  & 66.14  & 65.31  & 80.49  & 56.64  & 82.35  & 75.17  & 81.75  & 71.23  & 76.36 \\
            $+$ ML Lex.\ $+$ Panlex & 70.26  & 76.38  & 67.40  & 81.15  & 62.77  & 82.89  & 70.94  & 84.54  & 71.36  & 80.90 \\
            $+$ ML Lex.\ $+$ Panlex $+$ Filtering & 50.12  & 61.09  & 48.79  & 70.58  & 45.53  & 73.10  & 60.91  & 75.94  & 64.45  & 66.59 \\
            \hdashline
            mT5$_\text{Base}$ (580M) &  &  &  &  &  &  &  &  &  & \\
            $+$ fine-tuned on SST & 47.86  & 61.23  & 36.09  & 65.18  & 35.36  & 82.62  & 51.24  & 53.58  & 52.54  & 67.63 \\
            $+$ ML Lex.\ & 74.66  & 74.59  & 64.85  & 77.96  & 64.62  & 87.06  & 73.96  & 80.90  & 75.30  & 84.46 \\
            $+$ ML Lex.\ $+$ Panlex & 70.89  & 72.58  & 56.52  & 76.29  & 57.11  & 84.75  & 66.78  & 76.97  & 73.00  & 78.78 \\
            $+$ ML Lex.\ $+$ Panlex $+$ Filtering & 71.26  & 73.04  & 63.75  & 80.03  & 63.09  & 89.10  & 75.30  & 80.26  & 73.88  & 85.50 \\
            \hdashline
            mT5$_\text{Large}$ (1B) &  &  &  &  &  &  &  &  &  & \\
            $+$ fine-tuned on SST & 57.80  & 69.19  & 39.04  & 68.32  & 36.78  & 88.56  & 54.81  & 61.52  & 55.26  & 75.54 \\
            $+$ ML Lex.\ & 75.37  & 76.10  & 66.95  & 82.67  & 66.88  & 90.41  & 75.75  & 85.49  & 74.22  & 88.75 \\
            $+$ ML Lex.\ $+$ Panlex & 74.90  & 76.62  & 61.84  & 81.91  & 63.84  & 89.73  & 73.88  & 80.38  & 75.54  & 88.39 \\
            $+$ ML Lex.\ $+$ Panlex $+$ Filtering & 68.17  & 75.17  & 62.05  & 79.53  & 63.47  & 83.49  & 67.75  & 78.90  & 71.26  & 83.86 \\
        \bottomrule
        \end{tabular} 
    }
    \caption{All \textbf{binary 
    classification} results for NusaX.}
    \label{tab:b}
\end{table*}

\begin{table*}[ht]
    \centering
    \resizebox{\linewidth}{!}{
        \begin{tabular}{lccccccccccccccc}
        \toprule
            \multirow{2}{*}{\textbf{Method}} & \multicolumn{15}{c}{\textbf{African}} \\
             \cmidrule{2-16}
             & \textbf{am} & \textbf{dz} & \textbf{ha} & \textbf{ig} & \textbf{kr} & \textbf{ma} & \textbf{pcm} & \textbf{pt} & \textbf{sw} & \textbf{ts} & \textbf{twi} & \textbf{yo} & \textbf{or} & \textbf{tg} & \textbf{aeb} \\
             \midrule
                \multicolumn{3}{l}{\textbf{Full training}} \\
                mBERT$_\text{Base}$ & 67.44  & 70.15  & 88.09  & 91.23  & 76.25  & 67.91  & 75.10  & 77.52  & 73.39  & 63.60  & 78.48  & 83.17  & 54.96  & 49.18  & 75.37 \\
                $+$ ML Lex.\ & 65.50  & 71.95  & 88.89  & 91.74  & 79.96  & 69.86  & 75.83  & 78.70  & 77.37  & 64.71  & 77.67  & 84.11  & 51.76  & 49.12  & 75.93 \\
                $+$ ML Lex.\ $+$ Panlex  & 70.13  & 72.07  & 88.91  & 91.94  & 78.03  & 69.20  & 75.89  & 79.51  & 76.09  & 64.63  & 77.64  & 84.29  & 52.01  & 50.01  & 77.34 \\
                $+$ ML Lex.\ $+$ Panlex $+$ Filtering & 68.37  & 72.63  & 89.14  & 91.96  & 77.65  & 67.95  & 76.60  & 79.19  & 75.65  & 64.30  & 79.04  & 83.67  & 55.33  & 49.36  & 75.88 \\
            \midrule
                \multicolumn{3}{l}{\textbf{Training with limited data}} \\ 
                mBERT$_\text{Base}$ & 45.84  & 62.02  & 71.98  & 61.94  & 62.18  & 52.72  & 56.23  & 66.55  & 66.28  & 56.83  & 53.55  & 65.36  & 42.98  & 45.98  & 63.91 \\
                $+$ ML Lex.\ & 48.62  & 65.39  & 75.71  & 71.50  & 71.70  & 58.40  & 61.62  & 75.06  & 70.79  & 55.63  & 54.97  & 67.81  & 47.21  & 45.45  & 62.11 \\
                $+$ ML Lex.\ $+$ Panlex  & 50.92  & 64.31  & 74.09  & 70.68  & 70.96  & 58.19  & 62.81  & 75.67  & 73.27  & 54.86  & 56.71  & 67.91  & 52.80  & 45.67  & 64.08 \\
                $+$ ML Lex.\ $+$ Panlex $+$ Filtering & 53.07  & 65.11  & 75.05  & 71.41  & 69.94  & 57.91  & 63.26  & 76.17  & 70.74  & 55.26  & 56.39  & 68.23  & 53.60  & 45.24  & 63.73 \\
                \midrule
                \multicolumn{8}{l}{\textbf{Zero-shot (LLMs)}} \\ 
                XGLM (2.9B) & 25.28  & 30.34  & 45.48  & 41.24  & 45.14  & 52.24  & 50.75  & 47.25  & 56.46  & 43.94  & 43.08  & 39.23  & 38.70  & 33.57  & 44.78 \\
                BLOOMZ (3B) & 62.98  & 67.26  & 51.12  & 52.65  & 64.79  & 57.67  & 62.69  & 80.72  & 45.90  & 49.09  & 44.90  & 38.46  & 49.45  & 48.34  & 47.76 \\
                GPT--3.5 (174B) & 41.45  & 54.90  & 57.76  & 51.12  & 50.35  & 59.39  & 63.08  & 70.34  & 70.89  & 54.10  & 51.46  & 50.69  & 40.06  & 39.76  & 52.02 \\
                \midrule
                \multicolumn{8}{l}{\textbf{Zero-shot (SST and Lexicon-based pretraining with regression)}} \\
                mBERT$_\text{Base}$ (110M) &  &  &  &  &  &  &  &  &  &  &  &  &  &  & \\
                $+$ fine-tuned on SST & 31.75  & 59.08  & 53.21  & 47.76  & 46.52  & 55.87  & 61.74  & 65.13  & 32.51  & 37.70  & 40.13  & 31.84  & 49.33  & 37.10  & 48.71 \\
                $+$ ML Lex.\ & 51.10  & 61.20  & 61.06  & 67.50  & 62.64  & 56.61  & 65.10  & 69.06  & 72.03  & 52.79  & 45.46  & 58.49  & 48.53  & 47.10  & 48.24 \\
                $+$ ML Lex.\ $+$ Panlex & 52.56  & 61.67  & 60.16  & 69.96  & 65.34  & 60.17  & 66.72  & 72.27  & 68.01  & 46.52  & 50.21  & 60.31  & 49.92  & 48.75  & 52.36 \\
                $+$ ML Lex.\ $+$ Panlex $+$ Filtering & 50.66  & 61.24  & 53.87  & 68.89  & 58.36  & 58.04  & 67.41  & 72.45  & 68.46  & 53.33  & 51.20  & 62.08  & 47.30  & 48.33  & 48.45 \\
                \hdashline
                XLM-R$_\text{Base}$ (270M) &  &  &  &  &  &  &  &  &  &  &  &  &  &  & \\
                $+$ fine-tuned on SST & 80.76  & 72.71  & 65.89  & 45.38  & 48.62  & 60.71  & 65.70  & 83.67  & 59.49  & 39.69  & 38.23  & 27.30  & 48.04  & 60.13  & 52.57 \\
                $+$ ML Lex.\ & 81.62  & 68.67  & 60.54  & 64.43  & 59.22  & 59.36  & 64.97  & 80.31  & 70.30  & 56.90  & 51.12  & 34.80  & 53.18  & 62.36  & 54.03 \\
                $+$ ML Lex.\ $+$ Panlex & 81.17  & 70.19  & 60.81  & 65.92  & 63.80  & 61.00  & 66.05  & 79.58  & 72.92  & 60.03  & 52.09  & 40.16  & 54.50  & 65.09  & 53.13 \\
                $+$ ML Lex.\ $+$ Panlex $+$ Filtering & 81.14  & 70.54  & 64.42  & 65.06  & 60.92  & 61.08  & 67.56  & 80.31  & 75.35  & 58.16  & 51.96  & 57.85  & 54.33  & 62.04  & 49.19 \\
                \hdashline
                XLM-R$_\text{Large}$ (550M) &  &  &  &  &  &  &  &  &  &  &  &  &  &  & \\
                $+$ fine-tuned on SST & 82.92  & 75.06  & 70.90  & 52.31  & 56.65  & 64.66  & 66.81  & 85.71  & 69.60  & 46.03  & 44.20  & 36.17  & 55.45  & 62.47  & 57.34 \\
                $+$ ML Lex.\ & 81.52  & 69.43  & 71.28  & 70.63  & 62.06  & 64.88  & 67.72  & 80.68  & 77.75  & 61.23  & 53.34  & 53.72  & 53.95  & 67.05  & 57.19 \\
                $+$ ML Lex.\ $+$ Panlex & 78.26  & 69.28  & 68.08  & 67.55  & 52.02  & 60.45  & 68.38  & 83.45  & 80.70  & 63.39  & 56.32  & 60.13  & 46.16  & 46.27  & 57.38 \\
                $+$ ML Lex.\ $+$ Panlex $+$ Filtering & 78.84  & 67.74  & 67.81  & 63.83  & 46.45  & 61.22  & 69.81  & 84.49  & 83.26  & 55.39  & 51.16  & 59.35  & 47.43  & 47.74  & 49.62 \\
                \hdashline
                mBART$_\text{Large}$ (600M) &  &  &  &  &  &  &  &  &  &  &  &  &  &  & \\
                $+$ fine-tuned on SST & 65.74  & 71.64  & 66.99  & 52.74  & 54.24  & 62.94  & 67.80  & 82.91  & 61.32  & 54.90  & 50.94  & 37.19  & 53.62  & 48.43  & 62.95 \\
                $+$ ML Lex.\ & 63.40  & 69.53  & 64.93  & 65.09  & 62.53  & 61.31  & 70.38  & 83.21  & 82.08  & 54.91  & 56.44  & 56.14  & 48.10  & 47.62  & 57.61 \\
                $+$ ML Lex.\ $+$ Panlex & 63.23  & 69.43  & 67.95  & 70.72  & 65.20  & 60.86  & 68.68  & 83.48  & 77.31  & 52.93  & 50.99  & 53.53  & 52.49  & 47.25  & 57.14 \\
                $+$ ML Lex.\ $+$ Panlex $+$ Filtering & 58.19  & 62.84  & 50.86  & 59.26  & 49.83  & 58.42  & 65.55  & 82.66  & 83.68  & 58.04  & 53.25  & 59.62  & 38.75  & 46.71  & 46.58 \\
                \hdashline
                mT5$_\text{Base}$ (580M) &  &  &  &  &  &  &  &  &  &  &  &  &  &  & \\
                $+$ fine-tuned on SST & 79.97  & 67.21  & 68.15  & 54.55  & 66.64  & 59.14  & 65.70  & 79.85  & 56.31  & 36.35  & 34.85  & 30.46  & 48.40  & 59.71  & 50.38 \\
                $+$ ML Lex.\ & 78.20  & 68.65  & 78.16  & 67.04  & 75.52  & 67.93  & 69.81  & 80.59  & 80.32  & 54.40  & 51.52  & 49.12  & 56.30  & 72.56  & 61.63 \\
                $+$ ML Lex.\ $+$ Panlex & 78.62  & 69.08  & 76.56  & 69.00  & 74.35  & 68.08  & 68.53  & 79.27  & 77.16  & 59.46  & 54.96  & 42.08  & 56.20  & 72.07  & 56.71 \\
                $+$ ML Lex.\ $+$ Panlex $+$ Filtering & 77.31  & 68.49  & 75.92  & 67.16  & 75.97  & 67.01  & 70.45  & 82.65  & 81.84  & 54.67  & 54.64  & 52.40  & 53.69  & 70.58  & 60.08 \\
                \hdashline
                mT5$_\text{Large}$ (1B) &  &  &  &  &  &  &  &  &  &  &  &  &  &  & \\
                $+$ fine-tuned on SST & 81.45  & 69.14  & 66.54  & 54.94  & 73.30  & 59.49  & 65.43  & 81.83  & 59.14  & 39.07  & 36.09  & 32.45  & 49.32  & 64.23  & 47.66 \\
                $+$ ML Lex.\ & 83.09  & 73.54  & 77.57  & 70.64  & 79.14  & 67.58  & 69.18  & 82.78  & 80.85  & 58.42  & 55.06  & 48.31  & 56.27  & 74.45  & 62.26 \\
                $+$ ML Lex.\ $+$ Panlex & 81.20  & 70.21  & 79.67  & 72.68  & 75.91  & 67.07  & 70.03  & 82.14  & 82.74  & 59.93  & 59.97  & 51.87  & 57.53  & 78.40  & 61.33 \\
                $+$ ML Lex.\ $+$ Panlex $+$ Filtering & 79.09  & 70.35  & 75.20  & 69.27  & 71.35  & 66.87  & 68.95  & 80.86  & 78.85  & 59.81  & 56.89  & 58.52  & 56.20  & 71.62  & 56.74 \\          
        \bottomrule
        \end{tabular} 
    }
    \caption{All \textbf{binary 
    classification} results for the 14 African languages from SemEval 2023.}
    \label{tab:c}
\end{table*}

\begin{table*}[ht]
    \centering
    \resizebox{0.9\linewidth}{!}{
        \begin{tabular}{lcccccccccc}
        \toprule
            \multirow{2}{*}{\textbf{Method}} & \multicolumn{6}{c}{\textbf{High/Medium-Resource}} & & \multicolumn{3}{c}{\textbf{Code-switching}} \\
            \cmidrule{2-7} \cmidrule{9-11}
             & \textbf{en} & \textbf{ar} & \textbf{es} & \textbf{ru} & \textbf{id} & \textbf{ja} &  & \textbf{en-es} & \textbf{en-ml} & \textbf{en-ta} \\
            \midrule
            \multicolumn{3}{l}{\textbf{Full training}} \\ 
            mBERT$_\text{Base}$ & 72.65  & 64.15  & 62.21  & 80.05  & 85.91  & 83.04  &  & 63.58  & 88.69  & 79.83 \\
            $+$ ML Lex.\ & 72.78  & 64.68  & 64.11  & 81.10  & 85.73  & 83.12  &  & 65.42  & 88.98  & 80.86 \\
            $+$ ML Lex.\ $+$ Panlex  & 72.44  & 65.85  & 64.14  & 81.97  & 86.06  & 82.54  &  & 64.34  & 89.58  & 79.73 \\
            $+$ ML Lex.\ $+$ Panlex $+$ Filtering & 73.10  & 64.78  & 63.57  & 81.70  & 85.92  & 82.67  &  & 64.91  & 89.28  & 81.11 \\
            \midrule
            \multicolumn{3}{l}{\textbf{Training with limited data}} \\ 
            mBERT$_\text{Base}$ & 49.49  & 49.54  & 43.58  & 73.32  & 71.15  & 60.96  &  & 43.48  & 72.05  & 77.95 \\
            $+$ ML Lex.\ & 59.55  & 50.05  & 53.79  & 77.52  & 76.45  & 66.20  &  & 56.05  & 79.97  & 77.78 \\
            $+$ ML Lex.\ $+$ Panlex  & 59.32  & 51.96  & 55.65  & 78.20  & 78.21  & 66.51  &  & 54.40  & 77.83  & 78.03 \\
            $+$ ML Lex.\ $+$ Panlex $+$ Filtering & 60.03  & 52.15  & 57.32  & 78.35  & 77.15  & 68.46  &  & 51.96  & 79.34  & 77.98 \\
            \midrule
            \multicolumn{8}{l}{\textbf{Zero-shot (LLMs)}} \\  
            XGLM (2.9B) & 39.88  & 29.44  & 37.37  & 16.52  & 54.86  & 52.25  &  & 31.89  & 57.58  & 60.78 \\
            BLOOMZ (3B) & 71.12  & 48.66  & 52.04  & 6.93  & 80.57  & 53.95  &  & 34.20  & 31.12  & 42.23 \\
            GPT--3.5 (175B) & 67.61  & 55.56  & 60.56  & 83.35  & 66.58  & 72.22  &  & 58.97  & 42.85  & 49.42 \\
            \midrule
            \multicolumn{8}{l}{\textbf{Zero-shot (SST and Lexicon-based pretraining with classification)}} \\ 
            mBERT$_\text{Base}$ (110M) &  &  &  &  &  &  &  &  &  & \\
            $+$ fine-tuned on SST & 70.13  & 43.44  & 52.00  & 42.54  & 57.21  & 39.26  &  & 40.82  & 12.98  & 20.73 \\
            $+$ ML Lex.\ & 53.06  & 49.53  & 50.80  & 69.44  & 54.21  & 54.62  &  & 52.22  & 65.38  & 62.81 \\
            $+$ ML Lex.\ $+$ Panlex & 53.76  & 49.00  & 48.64  & 67.20  & 55.72  & 56.54  &  & 53.06  & 67.10  & 63.65 \\
            $+$ ML Lex.\ $+$ Panlex $+$ Filtering & 49.90  & 49.48  & 50.83  & 64.17  & 50.87  & 55.68  &  & 55.21  & 60.96  & 60.27 \\
            \hdashline
            XLM-R$_\text{Base}$ (270M) &  &  &  &  &  &  &  &  &  & \\
            $+$ fine-tuned on SST & 73.54  & 56.51  & 62.46  & 77.20  & 69.01  & 76.18  &  & 50.32  & 25.64  & 32.94 \\
            $+$ ML Lex.\ & 50.79  & 52.38  & 49.41  & 72.99  & 58.75  & 69.05  &  & 47.23  & 31.06  & 38.23 \\
            $+$ ML Lex.\ $+$ Panlex & 48.79  & 52.82  & 48.63  & 72.38  & 58.96  & 68.30  &  & 47.89  & 33.24  & 39.26 \\
            $+$ ML Lex.\ $+$ Panlex $+$ Filtering & 46.10  & 54.11  & 50.20  & 73.62  & 58.87  & 69.62  &  & 51.26  & 32.34  & 36.79 \\
            \hdashline
            XLM-R$_\text{Large}$ (550M) &  &  &  &  &  &  &  &  &  & \\
            $+$ fine-tuned on SST & 76.07  & 61.67  & 63.52  & 83.19  & 64.17  & 80.30  &  & 54.75  & 23.79  & 31.59 \\
            $+$ ML Lex.\ & 52.82  & 56.18  & 53.57  & 69.47  & 64.13  & 65.55  &  & 54.93  & 66.14  & 69.50 \\
            $+$ ML Lex.\ $+$ Panlex & 53.07  & 57.94  & 56.12  & 73.80  & 70.06  & 63.99  &  & 56.75  & 69.11  & 68.34 \\
            $+$ ML Lex.\ $+$ Panlex $+$ Filtering & 57.25  & 58.00  & 57.57  & 69.45  & 74.67  & 64.01  &  & 59.01  & 58.74  & 61.03 \\
            \hdashline
            mBART$_\text{Large}$ (600M) &  &  &  &  &  &  &  &  &  & \\
            $+$ fine-tuned on SST & 74.18  & 56.87  & 61.34  & 82.02  & 56.44  & 76.21  &  & 46.81  & 20.15  & 27.61 \\
            $+$ ML Lex.\ & 51.98  & 57.05  & 51.20  & 67.36  & 59.81  & 70.84  &  & 53.70  & 48.22  & 61.36 \\
            $+$ ML Lex.\ $+$ Panlex & 49.09  & 56.82  & 54.20  & 69.91  & 54.94  & 69.95  &  & 51.58  & 44.43  & 52.51 \\
            $+$ ML Lex.\ $+$ Panlex $+$ Filtering & 47.30  & 53.74  & 49.83  & 71.56  & 45.16  & 67.12  &  & 48.30  & 31.99  & 42.00 \\
            \hdashline
            mT5$_\text{Base}$ (580M) &  &  &  &  &  &  &  &  &  & \\
            $+$ fine-tuned on SST & 71.14  & 52.12  & 58.48  & 61.45  & 68.82  & 70.98  &  & 46.00  & 18.54  & 28.58 \\
            $+$ ML Lex.\ & 47.54  & 51.64  & 48.11  & 60.88  & 69.66  & 66.07  &  & 49.57  & 62.13  & 68.41 \\
            $+$ ML Lex.\ $+$ Panlex & 47.50  & 49.59  & 48.96  & 58.89  & 62.64  & 65.62  &  & 48.18  & 60.45  & 66.90 \\
            $+$ ML Lex.\ $+$ Panlex $+$ Filtering & 51.85  & 50.53  & 52.23  & 70.95  & 68.98  & 65.93  &  & 52.56  & 59.94  & 64.76 \\
            \hdashline
            mT5$_\text{Large}$ (1B) &  &  &  &  &  &  &  &  &  & \\
            $+$ fine-tuned on SST & 65.71  & 37.38  & 43.89  & 55.95  & 51.91  & 51.10  &  & 36.49  & 14.22  & 23.55 \\
            $+$ ML Lex.\ & 51.61  & 54.48  & 53.08  & 70.37  & 65.08  & 67.75  &  & 51.95  & 65.78  & 66.55 \\
            $+$ ML Lex.\ $+$ Panlex & 50.24  & 52.71  & 49.18  & 68.20  & 62.20  & 65.68  &  & 47.55  & 67.86  & 70.75 \\
            $+$ ML Lex.\ $+$ Panlex $+$ Filtering & 48.71  & 54.31  & 49.56  & 70.24  & 56.60  & 66.88  &  & 54.53  & 58.84  & 68.24 \\

        \bottomrule
        \end{tabular} 
    }
    \caption{All \textbf{3-way classification} results for high/medium-resource languages and code-switched texts.}
    \label{tab:a2}
\end{table*}

\begin{table*}[ht]
    \centering
    \resizebox{\linewidth}{!}{
        \begin{tabular}{lcccccccccc}
        \toprule
            \multirow{2}{*}{\textbf{Method}} & \multicolumn{10}{c}{\textbf{NusaX}} \\
             \cmidrule{2-11}
             & \textbf{ace} & \textbf{ban} & \textbf{bbc} & \textbf{bjn} & \textbf{bug} & \textbf{jav} & \textbf{mad} & \textbf{min} & \textbf{nij} & \textbf{sun}\\
            \midrule
            \multicolumn{3}{l}{\textbf{Full training}} \\ 
                mBERT$_\text{Base}$ & 75.67  & 76.09  & 71.71  & 75.94  & 74.64  & 78.37  & 72.88  & 76.60  & 73.40  & 76.61 \\
                $+$ ML Lex.\ & 76.74  & 75.69  & 75.08  & 78.54  & 75.41  & 81.07  & 72.39  & 81.21  & 74.05  & 79.40 \\
                $+$ ML Lex.\ $+$ Panlex  & 77.31  & 76.55  & 75.01  & 78.12  & 76.08  & 82.09  & 74.90  & 79.35  & 74.84  & 79.35 \\
                $+$ ML Lex.\ $+$ Panlex $+$ Filtering & 76.91  & 76.41  & 74.50  & 80.17  & 74.79  & 82.34  & 75.01  & 78.43  & 75.63  & 79.91 \\
                \midrule
                \multicolumn{3}{l}{\textbf{Training with limited data}} \\ 
                mBERT$_\text{Base}$ & 60.66  & 62.71  & 60.40  & 66.41  & 58.91  & 68.77  & 60.87  & 63.22  & 62.09  & 69.10 \\
                $+$ ML Lex.\ & 63.21  & 66.87  & 63.02  & 64.98  & 64.08  & 76.34  & 61.16  & 69.56  & 63.56  & 73.84 \\
                $+$ ML Lex.\ $+$ Panlex  & 64.86  & 67.43  & 65.86  & 69.42  & 65.02  & 77.92  & 64.16  & 70.73  & 65.30  & 72.20 \\
                $+$ ML Lex.\ $+$ Panlex $+$ Filtering & 65.51  & 66.93  & 65.17  & 68.82  & 64.26  & 78.03  & 63.11  & 70.27  & 66.74  & 72.67 \\
                \midrule
                \multicolumn{8}{l}{\textbf{Zero-shot (LLMs)}} \\   
                XGLM (2.9B) & 32.90  & 35.42  & 28.20  & 35.14  & 27.02  & 37.60  & 34.54  & 36.28  & 31.79  & 35.77 \\
                BLOOMZ (3B) & 51.91  & 51.68  & 39.36  & 57.28  & 34.49  & 56.56  & 47.45  & 54.08  & 48.91  & 47.15 \\
                GPT--3.5 (175B) & 49.19  & 48.96  & 33.09  & 59.82  & 26.38  & 63.74  & 45.65  & 59.10  & 44.42  & 54.63 \\
                \midrule
                \multicolumn{8}{l}{\textbf{Zero-shot (SST and Lexicon-based pretraining with classification)}} \\ 
                mBERT$_\text{Base}$ (110M) &  &  &  &  &  &  &  &  &  & \\
                $+$ fine-tuned on SST & 24.89  & 30.64  & 23.33  & 30.83  & 23.80  & 34.03  & 27.41  & 33.43  & 28.66  & 32.44 \\
                $+$ ML Lex.\ & 35.04  & 43.33  & 36.73  & 43.62  & 36.73  & 60.91  & 45.16  & 45.25  & 42.75  & 52.25 \\
                $+$ ML Lex.\ $+$ Panlex & 36.91  & 41.95  & 39.15  & 42.40  & 37.21  & 57.96  & 42.18  & 44.58  & 41.04  & 51.29 \\
                $+$ ML Lex.\ $+$ Panlex $+$ Filtering & 35.95  & 38.95  & 30.83  & 40.70  & 28.02  & 54.72  & 37.23  & 42.34  & 37.52  & 45.51 \\
                 \hdashline
                XLM-R$_\text{Base}$ (270M) &  &  &  &  &  &  &  &  &  & \\
                $+$ fine-tuned on SST & 34.36  & 40.39  & 27.88  & 50.40  & 25.19  & 62.25  & 37.43  & 48.13  & 41.43  & 45.96 \\
                $+$ ML Lex.\ & 29.57  & 35.27  & 13.99  & 43.52  & 13.58  & 52.47  & 28.47  & 47.01  & 25.54  & 49.11 \\
                $+$ ML Lex.\ $+$ Panlex & 29.78  & 35.42  & 15.25  & 46.90  & 14.94  & 50.94  & 28.83  & 45.31  & 28.23  & 47.82 \\
                $+$ ML Lex.\ $+$ Panlex $+$ Filtering & 32.11  & 33.78  & 13.26  & 44.18  & 10.57  & 52.70  & 32.69  & 45.18  & 22.25  & 46.56 \\
                 \hdashline
                XLM-R$_\text{Large}$ (550M) &  &  &  &  &  &  &  &  &  & \\
                $+$ fine-tuned on SST & 42.83  & 50.93  & 25.56  & 64.92  & 22.00  & 76.62  & 44.20  & 61.42  & 42.41  & 65.42 \\
                $+$ ML Lex.\ & 43.37  & 43.84  & 30.07  & 51.68  & 32.12  & 59.79  & 44.57  & 55.74  & 42.20  & 56.11 \\
                $+$ ML Lex.\ $+$ Panlex & 48.76  & 51.25  & 43.75  & 56.95  & 37.66  & 62.81  & 48.42  & 60.28  & 52.68  & 65.80 \\
                $+$ ML Lex.\ $+$ Panlex $+$ Filtering & 42.39  & 47.69  & 31.67  & 49.93  & 26.87  & 61.18  & 47.59  & 56.83  & 43.68  & 58.89 \\
                 \hdashline
                mBART$_\text{Large}$ (600M) &  &  &  &  &  &  &  &  &  & \\
                $+$ fine-tuned on SST & 25.81  & 33.64  & 17.25  & 46.62  & 13.31  & 46.02  & 30.52  & 44.44  & 32.60  & 27.76 \\
                $+$ ML Lex.\ & 36.09  & 41.12  & 31.95  & 52.44  & 27.32  & 50.21  & 43.97  & 48.85  & 40.15  & 46.70 \\
                $+$ ML Lex.\ $+$ Panlex & 31.24  & 38.97  & 26.54  & 44.26  & 22.03  & 43.78  & 34.47  & 42.80  & 33.77  & 39.76 \\
                $+$ ML Lex.\ $+$ Panlex $+$ Filtering & 28.73  & 34.82  & 17.64  & 38.36  & 15.81  & 40.94  & 27.59  & 38.22  & 27.60  & 35.31 \\
                 \hdashline
                mT5$_\text{Base}$ (580M) &  &  &  &  &  &  &  &  &  & \\
                $+$ fine-tuned on SST & 29.77  & 37.00  & 25.70  & 35.83  & 27.29  & 55.31  & 31.09  & 33.12  & 34.25  & 46.41 \\
                $+$ ML Lex.\ & 48.47  & 48.56  & 44.55  & 55.73  & 45.58  & 61.24  & 48.94  & 55.49  & 52.53  & 55.26 \\
                $+$ ML Lex.\ $+$ Panlex & 49.56  & 49.90  & 38.50  & 56.42  & 42.40  & 62.17  & 44.03  & 53.39  & 49.45  & 56.43 \\
                $+$ ML Lex.\ $+$ Panlex $+$ Filtering & 45.54  & 41.50  & 31.16  & 53.93  & 29.57  & 62.00  & 41.52  & 53.02  & 45.69  & 55.26 \\
                 \hdashline
                mT5$_\text{Large}$ (1B) &  &  &  &  &  &  &  &  &  & \\
                $+$ fine-tuned on SST & 29.32  & 32.70  & 28.89  & 32.90  & 28.10  & 37.39  & 29.24  & 32.32  & 30.95  & 34.90 \\
                $+$ ML Lex.\ & 48.28  & 48.30  & 41.65  & 58.03  & 40.26  & 62.78  & 53.14  & 55.63  & 52.47  & 57.88 \\
                $+$ ML Lex.\ $+$ Panlex & 54.32  & 52.35  & 43.84  & 59.30  & 47.76  & 61.94  & 49.85  & 58.63  & 54.95  & 56.65 \\
                $+$ ML Lex.\ $+$ Panlex $+$ Filtering & 41.64  & 48.12  & 33.72  & 51.50  & 34.45  & 59.36  & 42.19  & 54.99  & 44.15  & 53.62 \\
                
        \bottomrule
        \end{tabular} 
    }
    \caption{All \textbf{3-way classification} results for NusaX.}
    \label{tab:b2}
\end{table*}

\begin{table*}[ht]
    \centering
    \resizebox{\linewidth}{!}{
        \begin{tabular}{lccccccccccccccc}
        \toprule
            \multirow{2}{*}{\textbf{Method}} & \multicolumn{15}{c}{\textbf{African}} \\
             \cmidrule{2-16}
             & \textbf{am} & \textbf{dz} & \textbf{ha} & \textbf{ig} & \textbf{kr} & \textbf{ma} & \textbf{pcm} & \textbf{pt} & \textbf{sw} & \textbf{ts} & \textbf{twi} & \textbf{yo} & \textbf{or} & \textbf{tg} & \textbf{aeb} \\
             \midrule
             \multicolumn{3}{l}{\textbf{Full training}} \\ 
                mBERT$_\text{Base}$ & 17.51  & 58.38  & 75.27  & 77.73  & 56.41  & 48.21  & 62.93  & 63.52  & 55.04  & 49.61  & 65.01  & 71.30  & 35.38  & 36.89  & 71.59 \\
                $+$ ML Lex.\ & 10.10  & 59.99  & 75.98  & 78.93  & 59.23  & 48.37  & 63.66  & 65.08  & 57.18  & 51.66  & 64.59  & 71.77  & 32.82  & 38.22  & 71.32 \\
                $+$ ML Lex.\ $+$ Panlex  & 11.48  & 59.75  & 75.75  & 78.72  & 59.29  & 47.93  & 64.36  & 65.13  & 56.15  & 49.19  & 64.16  & 72.49  & 33.58  & 38.25  & 71.80 \\
                $+$ ML Lex.\ $+$ Panlex $+$ Filtering & 12.14  & 59.04  & 75.70  & 79.29  & 58.20  & 47.27  & 64.65  & 64.86  & 56.36  & 49.89  & 65.44  & 71.82  & 34.67  & 39.27  & 73.16 \\
                 \midrule
                \multicolumn{3}{l}{\textbf{Training with limited data}} \\ 
                mBERT$_\text{Base}$ & 5.42  & 49.46  & 53.66  & 39.71  & 40.48  & 31.72  & 50.61  & 53.60  & 47.82  & 39.86  & 42.32  & 42.56  & 29.36  & 32.06  & 59.45 \\
                $+$ ML Lex.\ & 5.99  & 52.00  & 54.24  & 53.02  & 40.34  & 38.82  & 55.19  & 59.76  & 47.38  & 40.34  & 41.82  & 42.25  & 30.48  & 35.70  & 60.20 \\
                $+$ ML Lex.\ $+$ Panlex  & 6.20  & 51.63  & 50.19  & 53.87  & 39.54  & 36.49  & 55.30  & 60.46  & 51.81  & 41.17  & 44.31  & 43.53  & 31.05  & 33.47  & 57.72 \\
                $+$ ML Lex.\ $+$ Panlex $+$ Filtering & 5.27  & 52.00  & 50.53  & 53.84  & 41.15  & 38.41  & 56.11  & 61.02  & 50.06  & 42.80  & 43.12  & 42.97  & 30.86  & 35.55  & 59.43 \\
                 \midrule
                \multicolumn{8}{l}{\textbf{Zero-shot (LLMs)}} \\ 
                XGLM (2.9B) & 16.23  & 22.74  & 26.25  & 18.00  & 24.12  & 28.74  & 43.38  & 17.63  & 15.25  & 35.90  & 35.80  & 22.19  & 17.03  & 20.51  & 41.98 \\
                BLOOMZ (3B) & 52.47  & 53.82  & 27.95  & 23.05  & 32.61  & 34.12  & 54.69  & 16.07  & 13.67  & 40.01  & 36.76  & 21.13  & 20.70  & 34.58  & 45.51 \\
                GPT--3.5 (175B) & 22.03  & 40.45  & 45.05  & 40.29  & 37.67  & 41.62  & 50.57  & 59.19  & 51.48  & 33.49  & 29.09  & 33.52  & 32.82  & 22.29  & 32.42 \\
                 \midrule
                \multicolumn{8}{l}{\textbf{Zero-shot (SST and Lexicon-based pretraining with classification)}} \\  
                mBERT$_\text{Base}$ (110M) &  &  &  &  &  &  &  &  &  &  &  &  &  &  & \\
                $+$ fine-tuned on SST & 16.16  & 32.77  & 25.12  & 31.83  & 28.94  & 33.93  & 38.70  & 44.91  & 39.26  & 15.71  & 11.76  & 22.95  & 32.94  & 17.67  & 24.35 \\
                $+$ ML Lex.\ & 9.39  & 45.13  & 36.55  & 38.75  & 38.72  & 40.17  & 46.68  & 45.93  & 34.88  & 31.97  & 35.11  & 40.21  & 31.49  & 15.24  & 35.97 \\
                $+$ ML Lex.\ $+$ Panlex & 7.58  & 40.85  & 36.55  & 37.76  & 40.61  & 39.75  & 45.65  & 47.51  & 35.26  & 36.60  & 34.69  & 42.31  & 32.71  & 12.94  & 36.14 \\
                $+$ ML Lex.\ $+$ Panlex $+$ Filtering & 5.64  & 39.13  & 39.13  & 44.50  & 34.07  & 38.94  & 42.71  & 52.01  & 40.50  & 24.53  & 32.75  & 37.72  & 29.35  & 9.81  & 30.54 \\
                 \hdashline
                XLM-R$_\text{Base}$ (270M) &  &  &  &  &  &  &  &  &  &  &  &  &  &  & \\
                $+$ fine-tuned on SST & 61.62  & 43.28  & 33.82  & 33.92  & 33.74  & 36.17  & 47.65  & 51.17  & 44.29  & 17.91  & 20.39  & 23.65  & 31.96  & 34.30  & 23.51 \\
                $+$ ML Lex.\ & 48.15  & 37.74  & 43.92  & 41.78  & 35.17  & 36.31  & 31.30  & 60.23  & 55.02  & 33.12  & 19.57  & 24.09  & 36.74  & 20.48  & 21.03 \\
                $+$ ML Lex.\ $+$ Panlex & 48.60  & 40.92  & 43.16  & 43.11  & 36.16  & 37.12  & 31.05  & 59.20  & 53.36  & 33.74  & 18.96  & 24.41  & 36.35  & 24.39  & 18.28 \\
                $+$ ML Lex.\ $+$ Panlex $+$ Filtering & 41.84  & 35.22  & 39.44  & 42.71  & 29.06  & 35.57  & 33.69  & 62.43  & 53.13  & 24.66  & 22.09  & 24.27  & 35.09  & 19.21  & 24.68 \\
                 \hdashline
                XLM-R$_\text{Large}$ (550M) &  &  &  &  &  &  &  &  &  &  &  &  &  &  & \\
                $+$ fine-tuned on SST & 60.78  & 49.27  & 41.82  & 33.61  & 32.37  & 42.48  & 45.67  & 55.20  & 50.55  & 20.05  & 18.49  & 23.05  & 35.20  & 27.51  & 24.76 \\
                $+$ ML Lex.\ & 52.71  & 42.98  & 50.26  & 46.10  & 38.54  & 41.44  & 49.93  & 43.45  & 46.11  & 40.24  & 40.36  & 38.47  & 28.53  & 32.91  & 35.60 \\
                $+$ ML Lex.\ $+$ Panlex & 58.72  & 49.15  & 45.26  & 32.79  & 38.65  & 40.92  & 52.81  & 46.26  & 45.54  & 47.24  & 44.14  & 32.83  & 28.07  & 33.10  & 41.52 \\
                $+$ ML Lex.\ $+$ Panlex $+$ Filtering & 61.22  & 49.35  & 50.09  & 49.17  & 39.86  & 41.90  & 48.18  & 44.12  & 50.91  & 42.49  & 37.08  & 33.39  & 35.47  & 33.81  & 30.98 \\
                 \hdashline
                mBART$_\text{Large}$ (600M) &  &  &  &  &  &  &  &  &  &  &  &  &  &  & \\
                $+$ fine-tuned on SST & 6.19  & 42.10  & 23.12  & 32.42  & 29.14  & 35.25  & 39.74  & 57.52  & 49.02  & 12.53  & 14.57  & 23.87  & 28.16  & 8.97  & 17.25 \\
                $+$ ML Lex.\ & 57.28  & 42.14  & 36.71  & 43.00  & 35.83  & 38.61  & 39.75  & 55.71  & 51.55  & 29.21  & 23.80  & 29.45  & 39.28  & 38.26  & 27.17 \\
                $+$ ML Lex.\ $+$ Panlex & 25.49  & 33.58  & 29.64  & 40.12  & 27.91  & 35.70  & 35.43  & 60.61  & 54.16  & 21.83  & 18.87  & 26.18  & 29.34  & 18.25  & 16.99 \\
                $+$ ML Lex.\ $+$ Panlex $+$ Filtering & 27.30  & 21.83  & 24.94  & 39.75  & 29.31  & 33.02  & 29.58  & 62.92  & 54.19  & 21.33  & 17.03  & 30.69  & 28.52  & 19.56  & 13.68 \\
                 \hdashline
                mT5$_\text{Base}$ (580M) &  &  &  &  &  &  &  &  &  &  &  &  &  &  & \\
                $+$ fine-tuned on SST & 57.69  & 44.85  & 40.66  & 34.80  & 36.91  & 40.86  & 50.48  & 44.64  & 38.68  & 23.48  & 23.66  & 22.65  & 31.24  & 38.38  & 34.94 \\
                $+$ ML Lex.\ & 60.93  & 50.01  & 48.75  & 41.54  & 45.58  & 44.56  & 46.89  & 44.97  & 36.91  & 38.82  & 39.60  & 27.89  & 28.33  & 51.10  & 52.30 \\
                $+$ ML Lex.\ $+$ Panlex & 62.41  & 51.19  & 47.82  & 43.07  & 44.17  & 44.95  & 53.56  & 48.19  & 42.02  & 33.94  & 40.23  & 26.34  & 29.17  & 52.87  & 49.41 \\
                $+$ ML Lex.\ $+$ Panlex $+$ Filtering & 63.52  & 50.95  & 43.89  & 45.86  & 44.18  & 44.97  & 51.92  & 48.81  & 48.98  & 34.52  & 34.56  & 34.42  & 31.24  & 49.92  & 38.98 \\
                 \hdashline
                mT5$_\text{Large}$ (1B) &  &  &  &  &  &  &  &  &  &  &  &  &  &  & \\
                $+$ fine-tuned on SST & 43.87  & 35.09  & 33.22  & 29.86  & 33.75  & 32.12  & 46.47  & 44.47  & 26.31  & 20.38  & 21.80  & 21.63  & 27.80  & 27.45  & 28.68 \\
                $+$ ML Lex.\ & 55.66  & 50.79  & 51.59  & 43.68  & 45.17  & 43.39  & 47.26  & 46.83  & 40.84  & 37.76  & 38.81  & 36.61  & 25.23  & 52.16  & 50.53 \\
                $+$ ML Lex.\ $+$ Panlex & 63.45  & 54.80  & 53.36  & 43.53  & 41.60  & 44.54  & 53.44  & 33.77  & 44.00  & 39.55  & 43.49  & 32.35  & 31.39  & 53.37  & 44.17 \\
                $+$ ML Lex.\ $+$ Panlex $+$ Filtering & 56.09  & 48.07  & 46.59  & 44.85  & 44.35  & 44.63  & 46.60  & 48.82  & 46.01  & 36.35  & 34.41  & 37.06  & 25.89  & 46.93  & 34.57 \\
                
        \bottomrule
        \end{tabular} 
    }
    \caption{All \textbf{3-way classification} results for the 14 African languages from SemEval 2023..}
    \label{tab:c2}
\end{table*}

\section{Additional Experiments}
\label{app:additional_data}

We present details of the datasets used in the additional experiments in Table~\ref{tab:data_additional}, and present the detailed results in Table~\ref{tab:result_additional}.

For binary 
experiment settings, we conduct the transformations:
\begin{itemize}
     \item \texttt{WT-WT} \cite{conforti-etal-2020-will}:: We consider the label \textit{support} as the positive class, \textit{refute} as the negative class, and discard \textit{comment} and \textit{unrelated} categories.
    \item \texttt{P-Stance} \cite{li-etal-2021-p}: We assign the label \textit{favor} as the positive class and \textit{against} as the negative class.
    \item \texttt{HS1} \cite{founta2018large}: We map the \textit{normal} class to the positive class and \textit{abusive} and \textit{hateful} classes to the negative class. We exclude the \textit{spam} class as it is unrelated to sentiment analysis.
    \item \texttt{HS2} \cite{vidgen-etal-2021-learning}: We map the \textit{none} label to the positive class and the remaining labels to the negative class. We acknowledge that this projection introduces noise as there is no absolute positive class available in this dataset.
    \item \texttt{GoEmotions} \cite{demszky-etal-2020-goemotions}: For the positive class, we include emotions such as \textit{admiration}, \textit{amusement}, \textit{approval}, \textit{caring}, \textit{curiosity}, \textit{desire}, \textit{excitement}, \textit{gratitude}, \textit{joy}, \textit{love}, \textit{optimism}, \textit{pride}, \textit{realization}, \textit{relief}, and \textit{surprise}. For the negative class, we include emotions such as \textit{annoyance}, \textit{confusion}, \textit{disappointment}, \textit{disapproval}, \textit{disgust}, \textit{embarrassment}, \textit{fear}, \textit{grief}, \textit{nervousness}, \textit{remorse}, and \textit{sadness}. Since \texttt{GoEmotions} is a multi-label dataset, we tally the positive and negative counts for each sentence and discard sentences with an equal count of positive and negative labels.
    \item \texttt{SemEval2018} \cite{mohammad-etal-2018-semeval}: For the positive class, we include emotions such as \textit{anticipation}, \textit{joy}, \textit{love}, \textit{optimism}, \textit{pessimism}, \textit{surprise}, and \textit{trust}. For the negative class, we include emotions such as \textit{anger}, \textit{disgust}, \textit{fear}, and \textit{sadness}. Similar to \texttt{GoEmotions}, since \texttt{SemEval2018} is a multi-label dataset, we employ the same strategy to determine the final class.                         
\end{itemize}

\begin{table*}[ht]
    \centering
    \resizebox{\linewidth}{!}{
        \begin{tabular}{llL{5cm}crr}
        \toprule
        \textbf{Task} & \textbf{Data} & \textbf{Label} & \textbf{Multilabel} & \textbf{Original (train/dev/test)} & \textbf{Binary (train/dev/test)} \\ 
        \midrule
        \multirow{2}{*}{Stance} & WT-WT \cite{conforti-etal-2020-will} & support, refute, comment, unrelated & No & 41027/5128/5129 & 8663/1070/1151 \\
         & P-Stance \citet{li-etal-2021-p} & favor, against & No & 17224/2193/2157 & 17224/2193/2157 \\
        \midrule
        \multirow{2}{*}{Hate speech} & HS1 \cite{founta2018large} & abusive, normal, hateful, spam & No & 79996/10000/10000 & 68803/8560/8603 \\
         & HS2 \citet{vidgen-etal-2021-learning} &  none, derogation, animosity, dehumanization, threatening, support  & No & 27256/3422/3356 & 27256/3422/3356 \\
        \midrule
        \multirow{2}{*}{Emotion} & GoEmotions \cite{demszky-etal-2020-goemotions} & 27 emotions & Yes & 43410/5426/5427 & 28264/3566/3551 \\
         & SemEval2018 \cite{mohammad-etal-2018-semeval} & 11 emotions & Yes & 6838/886/3259 & 5902/795/2846 \\
        \bottomrule
        \end{tabular}
    }
    \caption{Datasets used in our additional experiments.}
    \label{tab:data_additional}
\end{table*}

\begin{table*}[t]
    \centering
    \resizebox{0.8\linewidth}{!}{
        \begin{tabular}{lccccccccc}
        \toprule
            \multirow{2}{*}{\textbf{Model}} & \multicolumn{2}{c}{\textbf{Stance}} && \multicolumn{2}{c}{\textbf{Hate Speech}} && \multicolumn{2}{c}{\textbf{Emotion}} \\
            \cmidrule{2-3} \cmidrule{5-6} \cmidrule{8-9}
            & \textbf{WT-WT} & \textbf{P-Stance} && \textbf{HS1} & \textbf{HS2} && \textbf{GoEmotions} & \textbf{SemEval2018} \\
            \midrule
            \textit{Binary classes} \\
            mBERT$_\text{Base}$ & 79.88 &  60.66 && 83.72 & 54.49 && 57.45 & 60.46 \\
            $+$ EN Lex.\ &  \textbf{85.39} & \textbf{61.16} && \textbf{87.95} & \textbf{54.74} && \textbf{72.17} & \textbf{79.62} \\
            \midrule
            \textit{Original classes} \\
            mBERT$_\text{Base}$ & 44.43 & 60.66 && 69.45 & 43.26 && 14.20 & 17.57 \\
            $+$ EN Lex.\ &  \textbf{44.97} & \textbf{61.12} && \textbf{71.28} & \textbf{43.89} && \textbf{14.91 }& \textbf{32.93} \\

        \bottomrule
        \end{tabular}
    }
    \caption{Lexicon-based pretraining performance (macro-F1) in six other English semantic tasks. The results are based on the limited training data scenario.}
    \label{tab:result_additional}
\end{table*}

\end{document}